\documentclass[11pt]{article}

\usepackage[final]{acl}

\usepackage{times}
\usepackage{latexsym}

\usepackage[T1]{fontenc}

\usepackage[utf8]{inputenc}

\usepackage{microtype}

\usepackage{inconsolata}

\usepackage{graphicx}

\usepackage{graphicx}
\usepackage{booktabs}
\usepackage{amsmath}
\usepackage[table]{xcolor}
\usepackage{amssymb}
\usepackage{subfigure}

\title{Embedding-based In-Context Prompt Training for \\Enhancing LLMs as Text Encoders }

\author{
    Ailiang Lin$^{1}$, 
    Zhuoyun Li$^{2}$, 
    Keyu Mao$^{1}$,
    Kotaro Funakoshi$^{1}$,
    Manabu Okumura$^{1}$\\
    $^{1}$Institute of Science Tokyo \quad $^{2}$Tencent\\
\texttt{\{linailiang, maokeyu, funakoshi, oku\}@lr.first.iir.isct.ac.jp}\\ \texttt{earyli@tencent.com}
}

\begin{document}
\maketitle
\begin{abstract}
Large language models (LLMs) have been widely explored for embedding generation. While recent studies show that in-context learning (ICL) effectively enhances the representational capability of LLMs by prepending a few task-related demonstrations, it causes substantial token overhead due to the increased sequence length. In this work, we propose EPIC, a novel embedding-based in-context prompt training strategy that leverages ICL to generate high-quality embeddings while reducing computational burden during both training and inference. This approach replaces discrete text demonstrations with their corresponding continuous embeddings, which not only encourages the LLM to align semantically-related text pairs during contrastive learning, but also requires the model to interpret demonstration embeddings as part of the in-context prompt. Consequently, EPIC-trained models achieve excellent embedding performance both with or without in-context prompts at inference time. Comprehensive experiments demonstrate that our method establishes new state-of-the-art results on the MTEB benchmark, surpassing frontier models trained solely on publicly available retrieval data. Extensive ablation studies further validate the effectiveness and necessity of our mechanism.
\end{abstract}

\begin{figure}[ht]
    \centering
    \includegraphics[width=0.95\columnwidth]{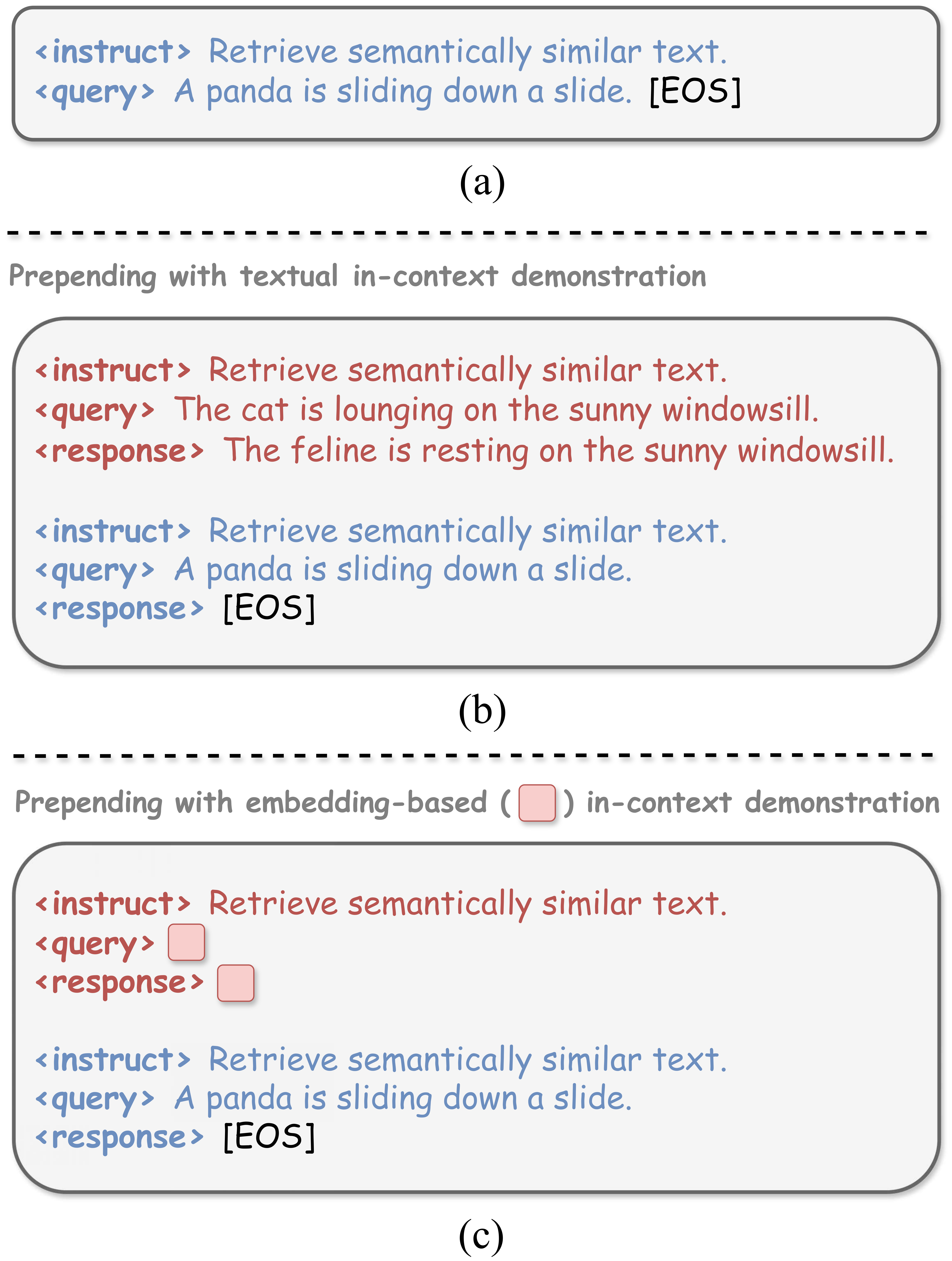}
    \caption{Comparison of different inputs for embedding tasks. (a) Embedding models typically take only the task instruction and user query as input. (b)~\citet{bgeicl} adopt the in-context learning strategy by incorporating task-related demonstrations. (c) EPIC enhances the input by prepending it with an embedding-based in-context prompt.}
    \label{fig:intro}
\end{figure}

\section{Introduction}
Text embeddings are powerful vector representations that capture contextual semantics of variable-length texts, playing a critical role in various natural language processing (NLP) tasks~\citep{mteb}. For example, retrieval-augmented generation (RAG) systems typically encode textual queries and documents into a shared embedding space, enabling efficient retrieval through similarity search~\citep{rag,liu2024chatqa}.

The rapid progress of Large Language Models (LLMs) brings new possibilities for improving the quality of text embeddings. Given the remarkable semantic understanding capabilities showcased by LLMs, recent research~\cite{grit,llm2vec,echo,nvembed,mgh,anchor} has increasingly focused on adapting them into text encoders through supervised contrastive learning~\cite{simcse,e5mistral}. 

In particular, PromptEOL~\cite{prompteol} incorporates in-context learning (ICL)~\citep{gpt3} into text embedding in a training-free manner. However,~\citet{grit} show that ICL cannot be directly applied to fine-tuned embedding models. To overcome this limitation, bge-en-icl~\cite{bgeicl} introduces a simple training strategy that effectively endows embedding models with ICL capabilities by prepending a few task-related \textit{query-passage} pairs (a.k.a. \textit{query–response} pairs) as demonstrations to the input text during contrastive learning. While these approaches highlight the potential of leveraging ICL to enhance text representation learning, their in-context demonstrations remain restricted to the discrete textual form, which substantially increases the input length and imposes a heavy token burden during training and inference, making them less practical in latency-sensitive scenarios, such as information retrieval and RAG tasks. Meanwhile, recent studies~\citep{icltaskvec,vectoricl} suggest that the ICL capabilities of LLMs can be extended to continuous vector representations under the next-token prediction paradigm, opening new avenues for more efficient exploitation of ICL.

In this context, we propose an \underline{\textbf{E}}mbedding-based \underline{\textbf{P}}rompt training with \underline{\textbf{I}}n-\underline{\textbf{C}}ontext demonstrations (\textbf{EPIC}), which leverages ICL to enhance the representational capability of LLMs while reducing computational overhead during both training and inference. Specifically, as shown in Figure~\ref{fig:intro}, we replace textual in-context demonstrations with their vector representations to form the embedding-based in-context prompt, which is then concatenated with the input query to obtain the desired query embedding. Since both the in-context and query embeddings are generated by the same model, contrastive learning not only encourages the LLM to align semantically-related positive pairs but also requires it to interpret the demonstration embeddings as part of the in-context prompt. During training, the demonstrations are directly sampled from the embeddings of positive pairs within the same batch. At inference time, we can pre-compute and reuse the embedding-based in-context prompts, avoiding redundant attention computation on textual demonstrations and thereby reducing inference latency.

We evaluate our EPIC on the Massive Text Embeddings Benchmark (MTEB)~\citep{mteb} across three popular LLMs, including Qwen2.5-7B, Mistral-7B, and LLaMA-3.1-8B. Experimental results show that our method achieves embedding performance on par with models trained with discrete textual ICL. Moreover, we observe an intriguing representational property: even without any in-context prompts during inference, the EPIC-trained models outperform the conventionally trained baselines under the same conditions. Notably, the proposed EPIC achieves new state-of-the-art results on MTEB among models trained exclusively on publicly available retrieval data. Extensive ablation studies further confirm the effectiveness and necessity of our approach. 

The primary contributions of this work are summarized as follows:
\begin{itemize}
\item We propose EPIC, a novel embedding-based in-context prompt training strategy that enhances LLMs as text encoders while reducing token overhead compared to textual ICL.

\item Experimental results demonstrate that LLMs trained with EPIC consistently improve embedding performance even without in-context demonstrations during inference.

\item EPIC-trained models achieve new state-of-the-art results on MTEB. We further provide in-depth ablation studies to validate the effectiveness and necessity of our method.
\end{itemize}

\section{Method}

\begin{figure*}[ht]
\centering
\includegraphics[width=\textwidth]{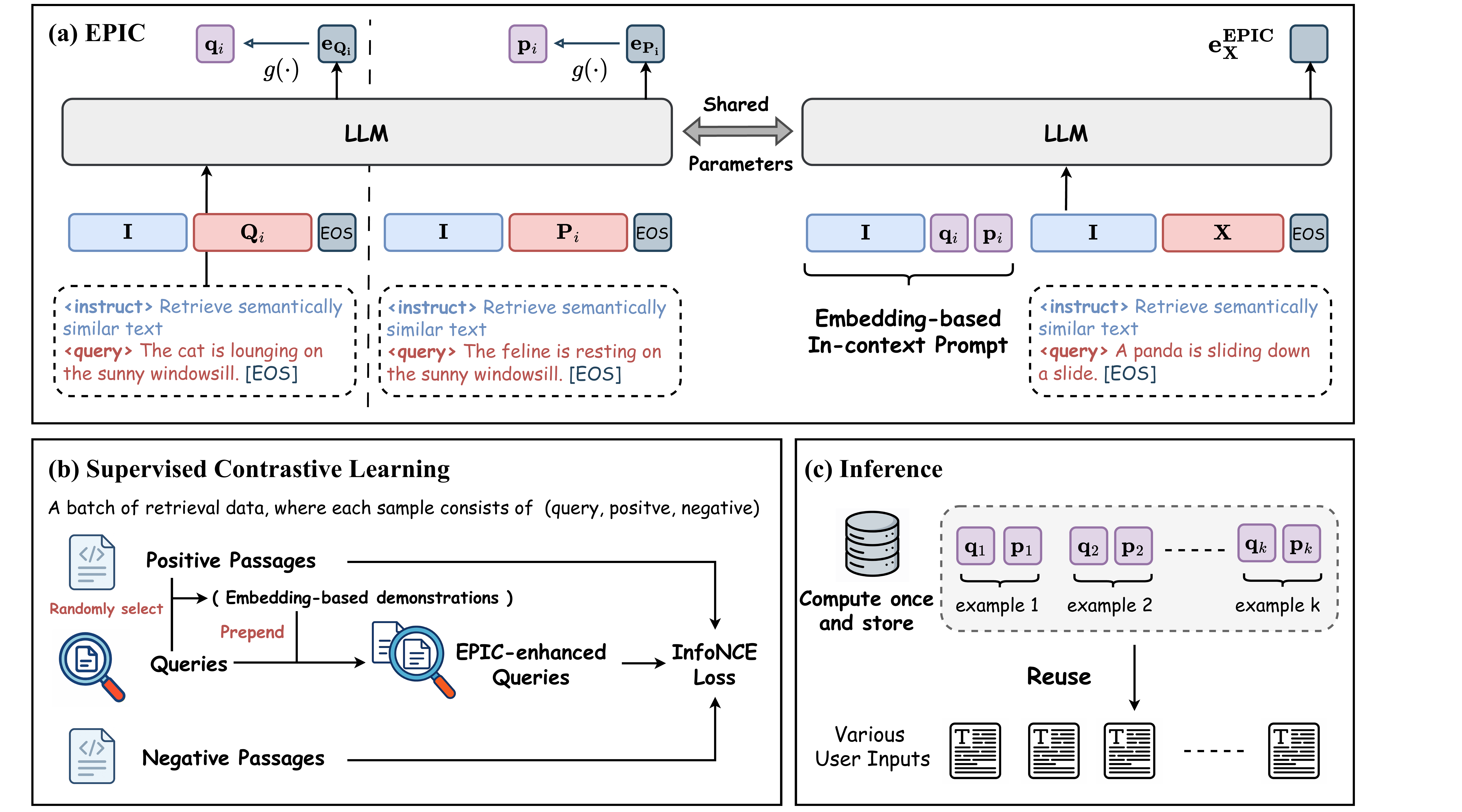}
\caption{{(a) Overview of the proposed EPIC method. For a given task (e.g., STS), the user input is "\textit{A panda is sliding down a slide}", while the demonstration query–passage pair consists of "\textit{The cat is lounging on the sunny windowsill}" and "\textit{The feline is resting on the sunny windowsill}". (b) During training, we randomly sample (\texttt{query}, \texttt{positive}) embedding pairs from the same batch as in-context demonstrations, which are then used to construct EPIC-enhanced \texttt{queries}. (c) The demonstration embeddings are pre-computed once and reused at inference time.}}
\label{fig:pipeline}
\end{figure*}

In this section, we first introduce the preliminaries of conventional in-context learning (ICL) for text embedding in Section~\ref{method:preliminary}. We then present our embedding-based in-context prompt (EPIC) method in Section~\ref{method:eicp}. Finally, we describe the training and inference strategies based on EPIC in Sections~\ref{method:training} and~\ref{method:inference}, respectively.

\subsection{Preliminary}
\label{method:preliminary}
For LLM-based embedding models, the text embedding is typically derived from the final hidden state of the special end-of-sequence (EOS) token, since only the last token can access the full sequence context under the causal attention mechanism. Specifically, given an input sequence $\mathbf{X}\in\mathbb{R}^{n\times d}$ of length $n$ with embedding dimension $d$, in addition to appending the $\texttt{[EOS]}$ token, we prepend a task-specific instruction $\mathbf{I}$, which enables the model to generalize across different embedding tasks~\cite{e5mistral}. The vector representation of the input text is formally defined as:
\begin{equation}
\mathbf{e}_\mathbf{X} = f_\theta^{\text{EOS}}([\mathbf{I}; \mathbf{X}; [\texttt{[EOS]}]])\in \mathbb{R}^d,
\end{equation}where 
$[\cdot; \cdot]$ denotes the sequence concatenation operation and $f_\theta^{\text{EOS}}(\cdot)$ refers to a function that returns the final hidden state of the LLM for the last input token, i.e., \texttt{[EOS]}.

Considering that the instruction alone provides limited information, bge-en-icl~\cite{bgeicl} expands the input sequence with a $k$-shot demonstration set $\mathcal{D} = \{\mathbf{D}_1, \mathbf{D}_2, \dots, \mathbf{D}_k\}$ to integrate the in-context learning (ICL) capabilities~\cite{gpt3} of LLMs into text embeddings. Concretely, each demonstration $\mathbf{D}_i$ consists of an instruction and a task-related \textit{query–passage} pair, i.e., $\mathbf{D}_i = [\mathbf{I}; \mathbf{Q}_i; \mathbf{P}_i]$,
as illustrated in Figure~\ref{fig:intro}(b). The ICL-based text embedding can be computed as:
\begin{equation}
\mathbf{e}^{\mathrm{ICL}}_\mathbf{X} = f_\theta^{\text{EOS}}([\mathbf{D}_1; \mathbf{D}_2;\dots; \mathbf{D}_k;\mathbf{I}; \mathbf{X}; [\texttt{[EOS]}]]).
\end{equation}

Notably, directly adding few-shot demonstrations in the prompts is generally ineffective for standard fine-tuned embedding models~\cite{grit}. Therefore, \textbf{\textit{the ICL in bge-en-icl and throughout the following discussion refers to capabilities acquired through specialized training strategies}}, rather than the original formulation without any gradient updates.

\subsection{Embedding-based In-Context Prompt}
\label{method:eicp}
While ICL has been shown to significantly enhance embedding quality~\cite{prompteol,bgeicl}, conventional in-context demonstrations introduce a large number of extra text tokens, leading to substantial computational overhead. This raises an intriguing question: \textit{could the embedding model benefit from ICL while mitigating the surge in sequence length}?

Inspired by the proven effectiveness of text embeddings, which inherently encode the contextual semantics of text, we challenge conventional wisdom by proposing an \textbf{E}mbedding-based \textbf{P}rompt training strategy with \textbf{I}n-\textbf{C}ontext demonstrations (\textbf{EPIC}) to improve the representational capacity of LLMs as text encoders. Specifically, as shown in Figure~\ref{fig:pipeline}(a), rather than using discrete textual demonstrations, we replace each \textit{query–passage} pair $(\mathbf{Q}_i, \mathbf{P}_i)$ with its corresponding continuous text embeddings. To further align these embedding-based demonstrations, we introduce a lightweight MLP layer $g(\cdot)$ consisting of two linear transformations with a GELU activation. The resulting continuous vector representations of the in-context \textit{query-passage} pair are computed as:
\begin{equation}
\begin{aligned}
\mathbf{q}_i &= g(f_\theta^{\text{EOS}}([\mathbf{I}; \mathbf{Q}_i; [\texttt{[EOS]}]]))\in \mathbb{R}^d, \\
\mathbf{p}_i &= g(f_\theta^{\text{EOS}}([\mathbf{I}; \mathbf{P}_i; [\texttt{[EOS]}]]))\in \mathbb{R}^d.
\end{aligned}
\end{equation}
The two vectors $\mathbf{q}_i$ and $\mathbf{p}_i$ compress the discrete \textit{query–passage} pair $(\mathbf{Q}_i, \mathbf{P}_i)$ into a shared latent space, substantially reducing token usage, since $|\mathbf{Q}_i| + |\mathbf{P}_i| \gg 2$, where $|\cdotp|$ denotes the sequence length. Accordingly, we transform the textual demonstration set $\mathcal{D}$ into an embedding-based version $\mathcal{E} = \{\mathbf{E}_1, \mathbf{E}_2, \dots, \mathbf{E}_k\}$, where each $\mathbf{E}_i = [\mathbf{I}; \mathbf{q}_i; \mathbf{p}_i]$. Consequently, the EPIC-enhanced embedding can be expressed as: 
\begin{equation}
\mathbf{e}^{\mathrm{EPIC}}_\mathbf{X} = f_\theta^{\text{EOS}}([\mathbf{E}_1; \mathbf{E}_2;\dots; \mathbf{E}_k; \mathbf{I}; \mathbf{X}; [\texttt{[EOS]}]]).
\end{equation}
Since the vector representations $\mathbf{q}_i$, $\mathbf{p}_i$, and $\mathbf{e}^{\mathrm{EPIC}}_\mathbf{X}$ all originate from the same LLM, which requires the model not only to generate high-quality embeddings but also to interpret its own embeddings when they are fed back as part of the in-context prompt. In this way, EPIC effectively reduces the token overhead of conventional ICL while preserving its representational advantages.

\subsection{Supervised Contrastive Learning}
\label{method:training}
In line with previous work~\cite{llm2vec, echo}, we fine-tune the LLM on publicly available retrieval datasets through contrastive learning, where each training sample consists of a triplet \texttt{(query, positive, negative)}. Consequently, each training step involves three forward passes to obtain the corresponding embeddings. To incorporate the proposed EPIC strategy, we perform an additional forward pass to generate the EPIC-enhanced \texttt{query} embedding (Figure~\ref{fig:pipeline}(b)). Following bge-en-icl~\cite{bgeicl}, we sample different \texttt{(query, positive)} embedding pairs from the same batch to construct the embedding-based in-context prompts, which are then used to enhance the original \texttt{Query}. The number of demonstration pairs is randomly chosen between 0 and a predefined maximum value, jointly enhancing the model’s representational capabilities with and without in-context prompts.

During training, we adopt the standard InfoNCE loss~\cite{loss}, defined as follows:
\begin{equation}
\mathcal{L} = -\log 
\frac{\phi(q, p^{+})}
{\phi(q, p^{+}) + \sum_{d^{-} \in \mathcal{N}} \phi(q, p^{-})},
\end{equation}
where $(q, p^{+})$ denotes the positive pair and $\mathcal{N}$ represents the set of in-batch and hard negative samples. The function $\phi(\cdot)$ is a temperature-scaled cosine similarity that measures the matching score between two text embeddings, computed as: 
\begin{equation}
\phi(q, p) = \exp (\frac{1}{\tau}\cos(\mathbf{e}_q, \mathbf{e}_p)),
\end{equation}
where $\tau$ is a temperature hyperparameter fixed to $0.05$ in our experiments.

\subsection{Inference}
\label{method:inference}
During inference, the proposed EPIC strategy may seem to increase computational cost since it requires generating additional vector representations. However, demonstration embeddings need to be computed only once, and the resulting embedding-based in-context prompt can be reused for the same task (Figure~\ref{fig:pipeline}(c)). This avoids repeatedly appending lengthy textual demonstrations at inference time, thereby reducing token usage while improving embedding quality. 

Furthermore, embedding performance under non-ICL settings is also crucial in practice. As discussed in Section~\ref{sec:main_res}, we observe a surprising representational effect: even without any in-context prompts during inference, the EPIC-trained models outperform the standard contrastive baselines under the same conditions. In contrast, models trained with conventional ICL do not exhibit such advantages when in-context demonstrations are removed, confirming the practicality of our EPIC. 

\section{Experiments}
\subsection{Experimental Setup}
\label{sect:experiment}
\paragraph{Training Datasets.} Following~\citet{llm2vec,bgeicl,mgh,anchor}, we conduct training on the public portion of the E5 dataset~\cite{e5mistral} curated by~\citet{echo}. The corpus is a collection of publicly available retrieval datasets, consisting of approximately 1.5M samples. Please refer to Appendix~\ref{appendix:training_data} for more details about the dataset composition.

\paragraph{Training Details.}
We apply the proposed EPIC to three popular LLMs: Qwen2.5-7B-Instruct (Qwen2.5-7B), Mistral-7B-Instruct-v0.2 (Mistral-7B), and Meta-Llama-3.1-8B-Instruct (LLaMA-3.1-8B). Following the training recipe from bge-en-icl~\cite{bgeicl}, we fine-tune the models using LoRA~\cite{hu2022lora} with rank 64, alpha 32, and a learning rate of $1e^{-4}$. For in-context demonstrations, we randomly sample 0 to 5 (\texttt{query}, \texttt{positive}) pairs from the in-batch training data. The maximum sequence length for training is set to 512 tokens. More training details are presented in Appendix~\ref{appendix:training_detail}.

\begin{table*}[t]
    \centering
    \small
    \resizebox{\textwidth}{!}{
    \tabcolsep=9pt
    \begin{tabular}{lcccccccc}
    \toprule
    \textbf{Categories $\rightarrow$} & \textbf{Retr.} & \textbf{Rerank.} & \textbf{Clust.} & \textbf{PairClass.} & \textbf{Class.} & \textbf{STS} & \textbf{Summ.} & \textbf{Avg} \\
    \textbf{\# of datasets $\rightarrow$} & \multicolumn{1}{c}{15}     & \multicolumn{1}{c}{4}     & \multicolumn{1}{c}{11}   & \multicolumn{1}{c}{3}      & \multicolumn{1}{c}{12}    & \multicolumn{1}{c}{10}  & \multicolumn{1}{c}{1}    & \multicolumn{1}{c}{56} \\ 
    \midrule
    \multicolumn{9}{c}{\textbf{\texttt{Miscellaneous}}}\\\midrule
    \multicolumn{1}{l}{SimCSE$_\text{BERT}$~\citep{simcse}} & 21.82 &47.54 & 33.43 & 73.68 & 67.32 & 79.12 & 23.31 & 48.72 \\
    \multicolumn{1}{l}{SGPT$_\text{5.8B}$~\citep{sgpt}} & 50.25 & 56.56 & 40.34 & 82.00 & 68.13 & 78.10 & 31.46 & 58.93\\
    \multicolumn{1}{l}{GTR$_\text{T5-XXL}$~\citep{gtr}} &48.48 & 56.65 & 42.42 & 86.12 & 67.41 & 78.38 & 30.64 & 58.97 \\
    \multicolumn{1}{l}{Sentence-T5$_\text{xxl}$~\citep{st5}}   &  42.24 & 56.42 & 43.72 & 85.07 & 73.42 & 82.63 & 30.08 & 59.51 \\
    \multicolumn{1}{l}{UDEVER$_\text{bloom-7b1}$~\citep{zhang2023language}} & 49.34 & 55.91 & 40.81 & 85.40 & 72.13 & 83.01 & 30.97 & 60.63 \\
    \multicolumn{1}{l}{Instructor$_\text{xl}$~\citep{su-etal-2023-one}} & 49.26 & 57.29 & 44.74 & 86.62 & 73.12 & 83.06 & \textbf{32.32} & 61.79\\
    \multicolumn{1}{l}{BGE$_\text{large-en-v1.5}$~\citep{xiao2024c}}   & 54.29 & 60.03 & 46.08 & 87.12 & 75.97 & 83.11 & 31.61 & 64.23 \\ 
    \multicolumn{1}{l}{UAE$_\text{large-v1}$~\citep{li-li-2024-aoe}}   & 54.66 & 59.88 & 46.73 & 87.25 & 75.58 & 84.54 & \underline{32.03} & 64.64 \\ 
    \midrule
    \multicolumn{9}{c}{\textbf{\texttt{Qwen2.5-7B}}}\\\midrule
    \rowcolor[gray]{0.9}EPIC (ours) &  56.52 & 59.53 & \underline{49.41} & 87.98 & 76.66 & 85.00 & 30.86 & 65.97 \\
    \midrule
    \multicolumn{9}{c}{\textbf{\texttt{LLaMA-3.1-8B}}}\\\midrule
    LLM2Vec~\citep{llm2vec} &  56.63 & 59.68 & 46.45 & 87.80 & 75.92 & 83.58 & 30.94 & 65.01 \\
    Anchor~\citep{anchor} &  57.09 & \textbf{61.38} & 46.03 & \textbf{88.92} & 76.17 & 83.76 & 30.13 & 65.30 \\
    \rowcolor[gray]{0.9}EPIC (ours) & 57.08 & 59.22 & 48.67 & 87.98 & 77.03 & \underline{85.38} & 31.26 & 66.10 \\
    \midrule
    \multicolumn{9}{c}{\textbf{\texttt{Mistral-7B}}}\\\midrule
    \multicolumn{1}{l}{E5~\citep{e5mistral}}  & 52.78 & 60.38 & 47.78 & 88.47 & 76.80 & 83.77 & 31.90 & 64.56 \\
    ECHO~\citep{echo} & 55.52 & 58.14 & 46.32 & 87.34 & 77.43 & 82.56 & 30.73 & 64.68\\
    GRITLM~\citep{grit} & 53.10 & \underline{61.30} & 48.90 & 86.90 & 77.00 & 82.80 & 29.40 & 64.70 \\
    LLM2Vec~\citep{llm2vec} & 55.99 & 58.42 & 45.54 & 87.99 & 76.63 & 84.09 & 29.96 & 64.80\\
    Anchor\citep{anchor} & 56.87 & 60.56 & 45.73 & 87.99 & 75.95 & 83.52 & 30.28 & 64.99 \\
    NV-Embed\dag~\citep{nvembed} & - & - & - & - & - & - & - & 65.80\\
    MGH~\citep{mgh} & \underline{57.49} & 58.80 & 47.96 & 87.83 & \textbf{77.62} & 84.04 & 31.10 & 65.87\\
    bge-en-icl~\citep{bgeicl} & \textbf{59.83} & 56.83 & 46.78 & 88.54 & \underline{77.51} & 84.08 & 30.39 & \underline{66.18}\\
    \rowcolor[gray]{0.9}EPIC (ours) &  56.89 & 59.52 & \textbf{49.56} & \underline{88.62} & 77.31 & \textbf{85.49} & 31.41 & \textbf{66.37} \\
    \midrule
    \end{tabular}
    }
    \caption{Performance comparison on the full MTEB benchmark (56 datasets) among models trained exclusively on publicly available retrieval data. \texttt{Qwen2.5-7B}, \texttt{Mistral-7B}, and \texttt{LLaMA-3.1-8B} denote models built upon these LLMs, while \texttt{Miscellaneous} refers to methods using other base models. \dag~represents the result is from~\citet{mgh}. The best result is highlighted in \textbf{bold}, and the second-best result is \underline{underlined}.}
    \label{tab:mteb}
\end{table*}

\paragraph{Evaluation.}
We verify the effectiveness of our method on the challenging Massive Text Embedding Benchmark (MTEB)~\cite{mteb}, which consists of 56 datasets spanning 7 diverse embedding tasks. Given that evaluating a 7B-parameter model on MTEB requires hundreds of A100 GPU hours, we conduct ablations and analysis on a smaller 26-dataset subset of MTEB. For fair comparison, we construct fixed in-context prompts for each dataset based on the examples provided by bge-en-icl. More evaluation details are presented in Appendix~\ref{appendix:evaluation}.

\begin{figure*}[t]
\centering
\includegraphics[width=1.\textwidth]{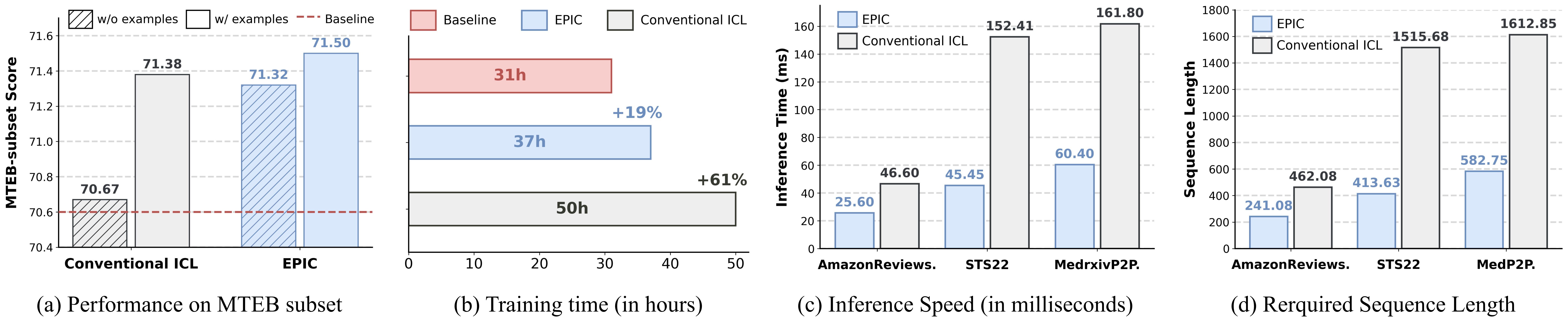}
\caption{{Comparison between EPIC and conventional ICL on Mistral-7B. (a) Performance comparison on the 26-dataset subset of MTEB  with and without in-context examples during inference. (b) Training time on a single NVIDIA A100 80GB GPU. (c) Average inference time per sample on selected MTEB datasets (see Appendix~\ref{ablation:inference} for more details). (d) Average required sequence length on selected MTEB datasets.}}
\label{fig:compared_icl}
\end{figure*}

\begin{table}[t]
\centering
\scalebox{0.77}{
\begin{tabular}{lccc}
\toprule
\textbf{Method} & \textbf{Qwen2.5-7B}  & \textbf{LLaMA-3.1-8B} & \textbf{Mistral-7B}\\
\midrule
Baseline & 65.05$_{{\mathbf{\phantom{+0.00}}}}$ & 65.25$_{{\mathbf{\phantom{+0.00}}}}$ & 65.33$_{{\mathbf{\phantom{+0.00}}}}$ \\
\rowcolor[gray]{0.9}EPIC$_{\text{w/o ICD}}$ & 65.68$_{{\mathbf{+0.63}}}$ & 65.89$_{{\mathbf{+0.64}}}$ & 66.11$_{{\mathbf{+0.78}}}$\\
\rowcolor[gray]{0.9}EPIC$_{\text{w/ ICD}}$ & \textbf{65.97}$_{{\mathbf{+0.92}}}$ & \textbf{66.10}$_{{\mathbf{+0.85}}}$ & \textbf{66.37}$_{{\mathbf{+1.04}}}$ \\
\bottomrule 
\end{tabular}
}
\caption{Performance of \textbf{EPIC-trained} models with or without in-context demonstrations (ICD) during inference on MTEB (56 datasets). Baseline models are conventionally trained without any ICL strategy.}
\label{tab:baseline}
\vspace{-0.2cm}
\end{table}

\subsection{Main Results}
\label{sec:main_res}
\paragraph{Comparison to state-of-the-art methods.} Since existing models~\cite{gemini,qwen3embed,zhao2025kalm} often rely on extensive in-domain non-retrieval data from MTEB or proprietary synthetic datasets for training, it is difficult to ensure a fair academic comparison and reliably assess generalization to unseen tasks~\cite{anchor,bgeicl}. To this end, we compare our EPIC only against models trained solely on publicly available retrieval datasets. 

Table~\ref{tab:mteb} presents the averaged scores for overall MTEB and its seven embedding task categories. Notably, our EPIC establishes new state-of-the-art performance across different LLM architectures. For the LLaMA-3.1-8B model, EPIC surpasses Anchor ($66.13$ vs. $65.30$), which requires an additional full-parameter training stage before contrastive learning. For the widely adopted Mistral-7B model, EPIC achieves an average score of $66.37$, outperforming E5 ($64.56$), ECHO ($64.68$), and bge-en-icl ($66.18$). Compared with bge-en-icl, which incorporates a conventional discrete ICL strategy, our findings suggest that embedding-based in-context prompting improves the representational capability more effectively. Moreover, EPIC exceeds competitive approaches that benefit from modified bidirectional attention on Mistral-7B, including GritLM ($64.70$), LLM2Vec ($64.80$), NV-Embed ($65.80$), and MGH ($65.87$). These results consistently showcase the superior performance of our EPIC in enhancing LLMs as text encoders across diverse embedding tasks.

\begin{figure}[t]
\centering
\includegraphics[width=0.5\textwidth]{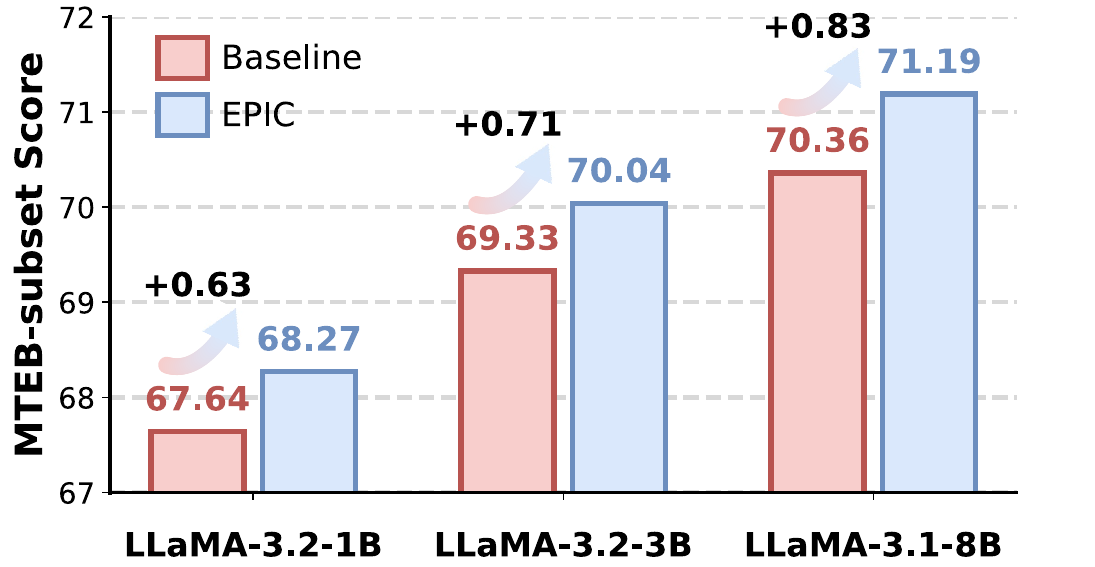}
\caption{Performance comparison on the MTEB subset across different model scales, including LLaMA-3.2-1B, LLaMA-3.2-3B, and LLaMA-3.1-8B.}
\label{fig:scale}
\end{figure}

\paragraph{Comparison to the baselines.}
In Table~\ref{tab:baseline}, we compare our EPIC (w/ or w/o in-context demonstrations during inference) against standard contrastive learning baselines that do not incorporate any ICL strategy. Specifically, our method yields notable performance improvements of $0.92$, $0.85$, and $1.04$ points over the baselines on Qwen2.5-7B, LLaMA-3.1-8B, and Mistral-7B, respectively. These results underscore the robustness and effectiveness of EPIC in improving embedding quality without relying on a specific base model.

Beyond improvements in in-context scenarios, we uncover an intriguing representational property: even without any in-context prompts at inference time, EPIC-trained models still achieve state-of-the-art performance, consistently outperforming baselines by $0.63$, $0.64$, and $0.78$ points on Qwen2.5-7B, LLaMA-3.1-8B, and Mistral-7B, respectively. We attribute this to three key factors during training: (1) the random sampling strategy explicitly allows the model to work without demonstrations; (2) the demonstration embeddings are generated without reliance on in-context prompts; and (3) EPIC not only encourages the model to align semantically related embeddings, but also requires it to internalize the demonstration embeddings as part of the in-context prompt.

Furthermore, compared to the baselines, EPIC-trained models consistently reduce the proportion of attention assigned to the first token across different layers, thereby alleviating the attention sink phenomenon~\cite{sink}. As a result, the EOS token is able to aggregate semantic information from the remaining tokens more effectively, leading to higher-quality embeddings. More details are provided in Appendix~\ref{appendix:sink}.

\begin{table}[t]
\centering
\scalebox{0.91}{
\begin{tabular}{p{2.7cm} cccc}
\toprule
\textbf{Sample-$n$} & $l$ & 64 & 32 & 16 \\
\midrule
Average Score & \textbf{71.50} & 71.40 & 71.32 & 71.30\\
\bottomrule 
\end{tabular}
}
\caption{{Performance of EPIC$_\text{Mistral-7B}$ on the MTEB subset by sampling $\frac{l}{n}$ tokens to represent the \textit{query} or \textit{passage} in demonstrations, where $l$ denotes the sequence length of $\mathbf{Q}_i$ or $\mathbf{P}_i$, and $n \in \{l, 64, 32, 16\}$.}}
\label{tab:ablation_number}
\end{table}

\paragraph{Comparison to discrete ICL.}
To further examine the benefits of our method, we quantitatively compare it against the Mistral-7B model trained with conventional ICL under the same settings. As illustrated in Figure~\ref{fig:compared_icl}(a), our continuous embedding-based strategy matches the performance of discrete textual ICL while requiring a lower token budget. More importantly, the ICL counterpart fails to improve embedding quality when demonstrations are removed, underscoring the superiority of our method in non-ICL scenarios.

Moreover, as shown in Figure~\ref{fig:compared_icl}(b), conventional ICL increases training time by over 60\% compared to the baseline, while EPIC incurs only about 19\% overhead by compressing discrete demonstrations into continuous vectors. In addition, Figure~\ref{fig:compared_icl}(c)-(d) confirm that EPIC consistently reduces token usage and yields lower inference latency on MTEB datasets, highlighting its efficiency in reducing computational cost during training and inference.

\subsection{Ablation Studies}
\label{sec:ablation}

\paragraph{Robustness across models of different scales.} Given the strong performance of EPIC on 7B and 8B models, we further evaluate its effectiveness at smaller scales. As shown in Figure~\ref{fig:scale}, EPIC consistently improves the embedding capabilities of LLMs ranging from 1B to 8B parameters, showing its scalability across model sizes. Furthermore, we observe larger gains as model size increases, indicating the potential for EPIC to continuously benefit from more powerful LLMs.

\begin{table}[t]
\centering
\scalebox{0.8}{
\begin{tabular}{l c}
\toprule
\textbf{Method} & \textbf{Average Score} \\
\midrule
\multicolumn{2}{c}{\textbf{(a) In-Context Prompt Format}}\\\midrule
w/o \textit{query}-\textit{passage} & 70.71 \\
w/o \textit{passage} & 71.17 \\
w/o \textit{query} & 71.31 \\
w/ only one instruction & 71.30 \\
\midrule
\multicolumn{2}{c}{\textbf{(b) Compression Strategy}}\\\midrule
Compress instruction and \textit{query-passage} & 71.22 \\
Only compress \textit{query} & 71.35 \\
Only compress \textit{passage}  &  71.42 \\\midrule
\rowcolor[gray]{0.9}EPIC (ours) &  \textbf{71.50} \\
\bottomrule 
\end{tabular}
}
\caption{Performance comparison of EPIC$_\text{Mistral-7B}$ on the MTEB subset with different in-context prompt formats and compression strategies.}
\label{tab:ablation_format_compress}

\end{table}
\begin{table}[t]
\centering
\scalebox{0.85}{
\begin{tabular}{p{5.7cm} c}
\toprule
\textbf{Method} & \textbf{Average Score} \\
\midrule
\rowcolor[gray]{0.9}EPIC &  \textbf{71.50} \\
\quad w/ Learnable tokens &  71.05 \\
Soft-Prompt & 70.83 \\
Instruction-Tuning &  70.60 \\
\bottomrule 
\end{tabular}
}
\caption{Performance comparison of EPIC$_\text{Mistral-7B}$ with other methods using learnable tokens on the MTEB subset, where Instruction-Tuning denotes the baseline trained using only task-specific instructions.}
\label{tab:ablation_soft}
\end{table}

\paragraph{The number of continuous vectors.} By default, EPIC uses two text embeddings to replace the  \textit{query}-\textit{passage} pair in discrete demonstrations. To examine whether using more continuous vectors could provide richer contextual information, we sample every $n$ tokens from the LLM's output sequence and represent the \textit{query} or \textit{passage} with $\frac{l}{n}$ continuous vectors (referred to as sample-$n$), where $l$ denotes the sequence length of $\mathbf{Q}_i$ or $\mathbf{P}_i$. Results for sample-$64/32/16$ are reported in Table~\ref{tab:ablation_number}. We observe that a single text embedding is sufficient for representing the \textit{query} or \textit{passage} in our setting, while increasing the number of continuous vectors does not yield performance improvements.

\paragraph{Impact of different in-context prompt formats.} The in-context demonstration used in this work consists of a textual instruction followed by a \textit{query}-\textit{passage} embedding pair. To examine the importance of this prompt design, we investigate four alternative prompt formats: (1) using only the instruction without \textit{query}-\textit{passage} embeddings, where each $\mathbf{E}_i = [\mathbf{I}]$; (2) retaining the instruction and the \textit{query} embedding while removing the \textit{passage} embedding, i.e., $\mathbf{E}_i = [\mathbf{I}; \mathbf{q}_i]$; (3) discarding only the \textit{query} embedding, i.e., $\mathbf{E}_i = [\mathbf{I}; \mathbf{p}_i]$; and (4) using only one instruction in the in-context prompt, yielding the input
$[\mathbf{I}; \mathbf{q}_1; \mathbf{p}_1; \dots; \mathbf{q}_k; \mathbf{p}_k; \mathbf{I}; \mathbf{X}; [\texttt{[EOS]}]]$. The results in Table~\ref{tab:ablation_format_compress}(a) indicate that all these variants lead to performance degradation, confirming the necessity of preserving the complete in-context prompt format adopted by EPIC.

\paragraph{Impact of different compression strategies.} In conventional ICL for embedding tasks~\cite{bgeicl}, each textual demonstration consists of an instruction and a \textit{query}-\textit{passage} pair. To challenge this paradigm, EPIC compresses both the discrete \textit{query} and \textit{passage} into their corresponding continuous embeddings. We further evaluate three alternative compression strategies: (1) transforming both the instruction and the \textit{query}-\textit{passage} pair into text embeddings; (2) compressing only the \textit{query}, and (3) compressing only the \textit{passage}. As demonstrated in Table~\ref{tab:ablation_format_compress}(b), EPIC exhibits the best trade-off between embedding performance and token usage. We hypothesize that jointly compressing the \textit{query}-\textit{passage} pair during training encourages the model to better understand and utilize its generated embeddings, while retaining the textual instruction effectively promotes the ICL capability.

\paragraph{Comparison with soft-prompt.} Since both soft prompts and our method fundamentally leverage continuous vectors to encode semantic information instead of hard prompts, we compare EPIC with two alternative setups to further highlight our contributions: (1) replacing the demonstration embeddings in EPIC with the same number of learnable tokens, and (2) following common practices~\cite{soft1,soft2} by prepending a set of learnable tokens as soft prompts to the input. All experiments are optimized with LoRA. The results in Table~\ref{tab:ablation_soft} show that EPIC achieves the best results, indicating that our embedding-based strategy provides richer semantic information in the continuous space than learnable tokens.

\begin{table}[t]
\centering
\scalebox{0.85}{
\begin{tabular}{p{5.7cm} c}
\toprule
\textbf{Method} & \textbf{Average Score} \\
\midrule
\rowcolor[gray]{0.9}EPIC &  \textbf{71.50} \\
\quad w/ Bi. + EOS pooling & 70.83 \\
\quad w/ Bi. + Mean pooling & 70.93 \\
\quad w/ Bi. + NV-Embed pooling & 71.08 \\
\bottomrule 
\end{tabular}
}
\caption{Performance of EPIC$_\text{Mistral-7B}$ on the MTEB subset using bidirectional (Bi.) attention with various pooling strategies. Note: EPIC preserves the original causal attention and employs EOS pooling by default.}
\label{tab:ablation_attention_pooling}
\end{table}
\paragraph{Influence of various attention and pooling mechanism.} Recent studies achieve strong text embeddings by transforming the model's attention from causal to bidirectional~\cite{bellm,grit,llm2vec,nvembed,mgh}. To investigate the potential of this paradigm in our framework, we evaluate EPIC under bidirectional attention with various pooling strategies, including last-token pooling, mean pooling, and NV-Embed pooling~\cite{nvembed}. As shown in Table~\ref{tab:ablation_attention_pooling}, we observe that switching to bidirectional attention considerably degrades EPIC's performance, regardless of the pooling mechanism, consistent with previous findings~\cite{bgeicl,lin2025causal2vec}. We speculate that the attention mismatch between pre-training and fine-tuning disrupts the advanced instruction-following capabilities of LLMs when provided with in-context demonstrations.

\section{Related Work}
\paragraph{Text Embeddings.} Text embeddings are continuous vector representations that encode the contextual semantics of natural language text, facilitating a wide range of natural language language processing (NLP) tasks such as text classification~\cite{class}, question answering~\cite{qa}, and information retrieval (IR)~\cite{cmedteb}. Early efforts focused on word-level embeddings~\cite{word1,word2}, while later attempts learned fixed-length representations for variable-length texts by combining word vectors~\cite{wieting2015towards,cse}. Modern approaches predominantly rely on pre-trained language models, such as BERT~\cite{bert}, RoBERTa~\cite{roberta}, and T5~\cite{t5} to generate contextualized text embeddings. Notable methods in this paradigm include SBERT~\cite{sbert}, SimCSE~\cite{simcse}, and Sentence-T5~\cite{st5}, which are fine-tuned on natural language inference datasets. To further improve embedding performance, advanced techniques such as E5~\cite{e5}, GTE~\cite{gte}, and BGE~\cite{xiao2024c} employ weakly supervised contrastive learning on large-scale text pair corpora curated from web sources. More recent work attempts to develop general-purpose embedding models tailored to diverse tasks and domains through well-designed instruction-tuning~\cite{instrut2,instrut3}.

\paragraph{LLM-based Text Embedding.} With the rapid advancement of large language models (LLMs), substantial efforts have been devoted to adapting them into strong embedding models. RepLLaMA~\cite{repllama} and LLaMA2Vec~\cite{llama2vec} show that fine-tuning LLaMA-2-7B~\cite{llama2} substantially improves the performances on retrieval tasks. To further obtain high-quality text embeddings, \citet{e5mistral}~fine-tune Mistral-7B~\cite{mistral} on diverse synthetic data with standard contrastive loss, achieving competitive results. ECHO~\cite{echo} repeats the input twice and extracts embeddings from the repeated sequence. Anchor~\cite{anchor} enhances the semantic capacity of the EOS token by introducing an additional training stage before contrastive learning. As the first work to enable bidirectional attention in LLMs for embedding generation, BeLLM~\cite{bellm} removes the causal mask at specific attention layers. Building on this foundation, many subsequent methods modify the LLMs to be fully bidirectional, including GRITLM~\cite{grit} and LLM2Vec~\cite{llm2vec}, while NV-Embed~\cite{nvembed} and MGH~\cite{mgh} further propose novel pooling strategies to overcome the limitation of mean pooling. In addition, PromptEOL~\citep{prompteol} and bge-en-icl~\citep{bgeicl} incorporate task-related demonstrations into the input to activate the in-context learning capabilities~\cite{gpt3} of LLMs. In this work, we aim to enhance LLMs as embedding models by leveraging ICL while mitigating its significant token cost through compressing discrete textual demonstrations into continuous embeddings. 

\paragraph{Vector-based ICL.} In-context learning (ICL) has become a powerful learning paradigm for LLMs, yet its underlying mechanisms remain unclear. \citet{icltaskvec} show that ICL operates by compressing a training set into a single task vector that guides the model to generate desired outputs. Building on this perspective, \citet{yang2025task} investigate potential factors in the emergence of task vectors. Moreover, \citet{zhuang2025vectoricl} demonstrate that pre-training projection modules with language modeling objectives enable effective vector-based ICL. Notably, these methods are developed for generative tasks. In contrast, to the best of our knowledge, this work presents the first embedding model that replaces discrete ICL demonstrations with their corresponding text embeddings, thus improving the representational capability of LLMs.

\section{Conclusion}
In this work, we introduced a novel embedding-based in-context prompt training strategy to improve the embedding capabilities of LLMs. Our method replaces conventional discrete demonstrations with their continuous embeddings, allowing the model to benefit from ICL while effectively reducing token overhead. Extensive experiments on MTEB demonstrated that EPIC achieves new state-of-the art results among models trained solely on publicly available retrieval datasets. Moreover, EPIC-enhanced models exhibited strong embedding performance even without any in-context prompt, further confirming the effectiveness and practicality of our method. We hope this work provides new perspective on prompting strategies for advancing the representation learning of LLMs.

\section*{Limitations}
Despite the strong embedding results achieved by EPIC, there remain several limitations that need to be acknowledged: (1) Models that perform exceptionally well on MTEB, such as Qwen3-Embedding~\cite{qwen3embed}, Gemini Embedding~\cite{gemini}, and KaLM-Embedding~\cite{zhao2025kalm}, typically rely on extensive synthetic or MTEB-related data during training. Incorporating such training corpora could help further validate the effectiveness and generalizability of our approach. (2) Due to hardware constraints, we evaluate the proposed method only on LLMs ranging from 1B to 8B parameters, which also ensures fair comparison with prior work~\cite{anchor,mgh,nvembed,bgeicl}. Scaling the experiments to larger model sizes, such as 30B or 70B, would make this work more comprehensive and meaningful. (3) Although this work provides new perspectives on embedding prompting, the underlying mechanisms of ICL for embedding generation remain unclear. Future work aims to provide a mechanistic explanation of ICL and further exploit its potential for text embedding.

\section*{Ethical Considerations}
This work focuses on improving LLMs as text encoders, enabling a wide range of real-world applications such as information retrieval, question answering, and recommendation systems. However, it should be noted that our method may inherit and potentially amplify social biases~\cite{hida-etal-2025-social} and hallucination issues~\cite{bang-etal-2025-hallulens} inherent in LLMs. Therefore, users are encouraged to apply our research in an ethical and responsible manner. In addition, we rely solely on publicly available datasets for training and open-source benchmarks for evaluation, both of which have been widely adopted in academic research, helping to mitigate ethical concerns to a certain extent.

\bibliography{main}

@inproceedings{rag,
  title={Retrieval-Augmented Generation for Knowledge-Intensive NLP Tasks},
  author={Lewis, Patrick and Perez, Ethan and Piktus, Aleksandra and Petroni, Fabio and Karpukhin, Vladimir and Goyal, Naman and K{\"u}ttler, Heinrich and Lewis, Mike and Yih, Wen-tau and Rockt{\"a}schel, Tim and Riedel, Sebastian and Kiela Douwe},
  booktitle={Advances in Neural Information Processing Systems},
  volume={33},
  pages={9459--9474},
  year={2020}
}

@inproceedings{mteb,
  title={MTEB: Massive text embedding benchmark},
  author={Muennighoff, Niklas and Tazi, Nouamane and Magne, Lo{\"\i}c and Reimers, Nils},
  booktitle={Proceedings of the 17th Conference of the European Chapter of the Association for Computational Linguistics},
  pages={2014--2037},
  year={2023}
}

@inproceedings{bert,
  title={Bert: Pre-training of deep bidirectional transformers for language understanding},
  author={Devlin, Jacob and Chang, Ming-Wei and Lee, Kenton and Toutanova, Kristina},
  booktitle={Proceedings of the 2019 conference of the North American chapter of the association for computational linguistics: human language technologies, volume 1 (long and short papers)},
  pages={4171--4186},
  year={2019}
}

@article{t5,
  title={Exploring the limits of transfer learning with a unified text-to-text transformer},
  author={Raffel, Colin and Shazeer, Noam and Roberts, Adam and Lee, Katherine and Narang, Sharan and Matena, Michael and Zhou, Yanqi and Li, Wei and Liu, Peter J},
  journal={Journal of Machine Learning Research},
  volume={21},
  number={140},
  pages={1--67},
  year={2020}
}

@inproceedings{karpukhin2020dense,
  title={Dense Passage Retrieval for Open-Domain Question Answering.},
  author={Karpukhin, Vladimir and Oguz, Barlas and Min, Sewon and Lewis, Patrick and Wu, Ledell and Edunov, Sergey and Chen, Danqi and Yih, Wen-tau},
  booktitle={Proceedings of the 2020 Conference on Empirical Methods in Natural Language Processing (EMNLP)},
  pages={6769--6781},
  year={2020}
}

@inproceedings{liu2024chatqa,
 title={Chat{QA}: Surpassing {GPT}-4 on Conversational {QA} and {RAG}},
 author={Zihan Liu and Wei Ping and Rajarshi Roy and Peng Xu and Chankyu Lee and Mohammad Shoeybi and Bryan Catanzaro},
 booktitle = {Advances in Neural Information Processing Systems},
 year={2024}
}

@inproceedings{llm2vec,
 title={{LLM}2Vec: Large Language Models Are Secretly Powerful Text Encoders},
 author={Parishad BehnamGhader and Vaibhav Adlakha and Marius Mosbach and Dzmitry Bahdanau and Nicolas Chapados and Siva Reddy},
 booktitle={First Conference on Language Modeling},
 year={2024}
}

@inproceedings{nvembed,
 title={{NV}-Embed: Improved Techniques for Training {LLM}s as Generalist Embedding Models},
 author={Chankyu Lee and Rajarshi Roy and Mengyao Xu and Jonathan Raiman and Mohammad Shoeybi and Bryan Catanzaro and Wei Ping},
 booktitle={The Thirteenth International Conference on Learning Representations},
 year={2025}
}

@inproceedings{grit,
  title={Generative Representational Instruction Tuning},
  author={Muennighoff, Niklas and Hongjin, SU and Wang, Liang and Yang, Nan and Wei, Furu and Yu, Tao and Singh, Amanpreet and Kiela, Douwe},
  booktitle={ICLR 2024 Workshop: How Far Are We From AGI},
  year={2024}
}

@inproceedings{echo,
title={Repetition Improves Language Model Embeddings},
author={Jacob Mitchell Springer and Suhas Kotha and Daniel Fried and Graham Neubig and Aditi Raghunathan},
booktitle={The Thirteenth International Conference on Learning Representations},
year={2025}
}

@inproceedings{bgeicl,
title={Making Text Embedders Few-Shot Learners},
author={Chaofan Li and Minghao Qin and Shitao Xiao and Jianlyu Chen and Kun Luo and Defu Lian and Yingxia Shao and Zheng Liu},
booktitle={The Thirteenth International Conference on Learning Representations},
year={2025}
}

@article{roberta,
  title={Roberta: A robustly optimized bert pretraining approach},
  author={Liu, Yinhan and Ott, Myle and Goyal, Naman and Du, Jingfei and Joshi, Mandar and Chen, Danqi and Levy, Omer and Lewis, Mike and Zettlemoyer, Luke and Stoyanov, Veselin},
  journal={arXiv preprint arXiv:1907.11692},
  year={2019}
}

@inproceedings{simcse,
    title = "{S}im{CSE}: Simple Contrastive Learning of Sentence Embeddings",
    author = "Gao, Tianyu  and
      Yao, Xingcheng  and
      Chen, Danqi",
    booktitle = "Proceedings of the 2021 Conference on Empirical Methods in Natural Language Processing",
    year = "2021",
    pages = "6894--6910"
}

@inproceedings{st5,
    title = "Sentence-T5: Scalable Sentence Encoders from Pre-trained Text-to-Text Models",
    author = "Ni, Jianmo  and
      Hernández Ábrego, Gustavo  and
      Constant, Noah  and
      Ma, Ji  and
      Hall, Keith B.  and
      Cer, Daniel  and
      Yang, Yinfei",
    booktitle = "Findings of the Association for Computational Linguistics: ACL 2022",
    year = "2022",
    pages = "1864--1874"
}

@article{e5,
  title={Text Embeddings by Weakly-Supervised Contrastive Pre-training},
  author={Wang, Liang and Yang, Nan and Huang, Xiaolong and Jiao, Binxing and Yang, Linjun and Jiang, Daxin and Majumder, Rangan and Wei, Furu},
  journal={arXiv preprint arXiv:2212.03533},
  year={2022}
}

@article{gte,
  title={Towards general text embeddings with multi-stage contrastive learning},
  author={Li, Zehan and Zhang, Xin and Zhang, Yanzhao and Long, Dingkun and Xie, Pengjun and Zhang, Meishan},
  journal={arXiv preprint arXiv:2308.03281},
  year={2023}
}

@inproceedings{repllama,
  title={Fine-tuning llama for multi-stage text retrieval},
  author={Ma, Xueguang and Wang, Liang and Yang, Nan and Wei, Furu and Lin, Jimmy},
  booktitle={Proceedings of the 47th International ACM SIGIR Conference on Research and Development in Information Retrieval},
  pages={2421--2425},
  year={2024}
}

@inproceedings{llama2vec,
    title = "{L}lama2{V}ec: Unsupervised Adaptation of Large Language Models for Dense Retrieval",
    author = "Liu, Zheng  and
      Li, Chaofan  and
      Xiao, Shitao  and
      Shao, Yingxia  and
      Lian, Defu",
    booktitle = "Proceedings of the 62nd Annual Meeting of the Association for Computational Linguistics (Volume 1: Long Papers)",
    year = "2024",
    pages = "3490--3500"
}

@inproceedings{prompteol,
    title = "Scaling Sentence Embeddings with Large Language Models",
    author = "Jiang, Ting  and
      Huang, Shaohan  and
      Luan, Zhongzhi  and
      Wang, Deqing  and
      Zhuang, Fuzhen",
    booktitle = "Findings of the Association for Computational Linguistics: EMNLP 2024",
    year = "2024",
    pages = "3182--3196"
}

@inproceedings{e5mistral,
    title = "Improving Text Embeddings with Large Language Models",
    author = "Wang, Liang  and
      Yang, Nan  and
      Huang, Xiaolong  and
      Yang, Linjun  and
      Majumder, Rangan  and
      Wei, Furu",
    booktitle = "Proceedings of the 62nd Annual Meeting of the Association for Computational Linguistics",
    year = "2024",
    pages = "11897--11916"
}

@article{loss,
  title={Unsupervised dense information retrieval with contrastive learning},
  author={Izacard, Gautier and Caron, Mathilde and Hosseini, Lucas and Riedel, Sebastian and Bojanowski, Piotr and Joulin, Armand and Grave, Edouard},
  journal={arXiv preprint arXiv:2112.09118},
  year={2021}
}

@article{llama2,
  title={Llama 2: Open foundation and fine-tuned chat models},
  author={Touvron, Hugo and Martin, Louis and Stone, Kevin and Albert, Peter and Almahairi, Amjad and Babaei, Yasmine and Bashlykov, Nikolay and Batra, Soumya and Bhargava, Prajjwal and Bhosale, Shruti and others},
  journal={arXiv preprint arXiv:2307.09288},
  year={2023}
}

@article{mistral,
  title={Mistral 7B},
  author={Jiang, Albert Q and Sablayrolles, Alexandre and Mensch, Arthur and Bamford, Chris and Chaplot, Devendra Singh and Casas, Diego de las and Bressand, Florian and Lengyel, Gianna and Lample, Guillaume and Saulnier, Lucile and others},
  journal={arXiv preprint arXiv:2310.06825},
  year={2023}
}

@inproceedings{hu2022lora,
title={Lo{RA}: Low-Rank Adaptation of Large Language Models},
author={Edward J Hu and yelong shen and Phillip Wallis and Zeyuan Allen-Zhu and Yuanzhi Li and Shean Wang and Lu Wang and Weizhu Chen},
booktitle={International Conference on Learning Representations},
year={2022},
url={https://openreview.net/forum?id=nZeVKeeFYf9}
}

@inproceedings{flashattention,
title={FlashAttention-2: Faster Attention with Better Parallelism and Work Partitioning},
author={Tri Dao},
booktitle={The Twelfth International Conference on Learning Representations},
year={2024},
url={https://openreview.net/forum?id=mZn2Xyh9Ec}
}

@inproceedings{eli5,
    title = "{ELI}5: Long Form Question Answering",
    author = "Fan, Angela  and
      Jernite, Yacine  and
      Perez, Ethan  and
      Grangier, David  and
      
      Weston, Jason  and
      Auli, Michael",
    booktitle = "Proceedings of the 57th Annual Meeting of the Association for Computational Linguistics",
    year = "2019",
    pages = "3558--3567"
}

@inproceedings{hotpotqa,
    title = "{H}otpot{QA}: A Dataset for Diverse, Explainable Multi-hop Question Answering",
    author = "Yang, Zhilin  and
      Qi, Peng  and
      Zhang, Saizheng  and
      Bengio, Yoshua  and
      Cohen, William W.  and
      Salakhutdinov, Ruslan  and
      Manning, Christopher D.",
    booktitle = "Proceedings of the 2018 Conference on Empirical Methods in Natural Language Processing",
    year = "2018",
    pages = "2369--2380"
}

@inproceedings{fever,
    title = "{FEVER}: a Large-scale Dataset for Fact Extraction and {VER}ification",
    author = "Thorne, James  and
      Vlachos, Andreas  and
      Christodoulopoulos, Christos  and
      Mittal, Arpit",
    booktitle = "Proceedings of the 2018 Conference of the North {A}merican Chapter of the Association for Computational Linguistics: Human Language Technologies, Volume 1 (Long Papers)",
    year = "2018",
    pages = "809--819"
}

@article{miracl,
    title = "{MIRACL}: A Multilingual Retrieval Dataset Covering 18 Diverse Languages",
    author = "Zhang, Xinyu  and
      Thakur, Nandan  and
      Ogundepo, Odunayo  and
      Kamalloo, Ehsan  and
      Alfonso-Hermelo, David  and
      Li, Xiaoguang  and
      Liu, Qun  and
      Rezagholizadeh, Mehdi  and
      Lin, Jimmy",
    journal = "Transactions of the Association for Computational Linguistics",
    volume = "11",
    year = "2023",
    pages = "1114--1131"
}

@misc{marco,
title={{MS} {MARCO}: A Human-Generated {MA}chine Reading {CO}mprehension Dataset},
author={Tri Nguyen and Mir Rosenberg and Xia Song and Jianfeng Gao and Saurabh Tiwary and Rangan Majumder and Li Deng},
year={2017},
url={https://openreview.net/forum?id=Hk1iOLcle}
}

@inproceedings{MrTyDi,
    title = "Mr. {T}y{D}i: A Multi-lingual Benchmark for Dense Retrieval",
    author = "Zhang, Xinyu  and
      Ma, Xueguang  and
      Shi, Peng  and
      Lin, Jimmy",
    booktitle = "Proceedings of the 1st Workshop on Multilingual Representation Learning",
    year = "2021",
    pages = "127--137"
}

@inproceedings{dureader,
    title = "{D}u{R}eader: a {C}hinese Machine Reading Comprehension Dataset from Real-world Applications",
    author = "He, Wei  and
      Liu, Kai  and
      Liu, Jing  and
      Lyu, Yajuan  and
      Zhao, Shiqi  and
      Xiao, Xinyan  and
      Liu, Yuan  and
      Wang, Yizhong  and
      Wu, Hua  and
      She, Qiaoqiao  and
      Liu, Xuan  and
      Wu, Tian  and
      Wang, Haifeng",
    booktitle = "Proceedings of the Workshop on Machine Reading for Question Answering",
    year = "2018",
    publisher = "Association for Computational Linguistics",
    url = "https://aclanthology.org/W18-2605/",
    pages = "37--46"
}

@inproceedings{t2ranking,
  title={T2ranking: A large-scale chinese benchmark for passage ranking},
  author={Xie, Xiaohui and Dong, Qian and Wang, Bingning and Lv, Feiyang and Yao, Ting and Gan, Weinan and Wu, Zhijing and Li, Xiangsheng and Li, Haitao and Liu, Yiqun and Ma, Jin},
  booktitle={Proceedings of the 46th International ACM SIGIR Conference on Research and Development in Information Retrieval},
  pages={2681--2690},
  year={2023}
}

@article{quora,
    author = {DataCanary and hilfialkaff and Jiang, Lili and Risdal, Meg and Dandekar, Nikhil and tomtung},
    title = {Quora Question Pairs},
    year = {2017},
    url = {https://kaggle.com/competitions/quora-question-pairs}
}

@inproceedings{squad,
    title = "{SQ}u{AD}: 100,000+ Questions for Machine Comprehension of Text",
    author = "Rajpurkar, Pranav  and
      Zhang, Jian  and
      Lopyrev, Konstantin  and
      Liang, Percy",
    booktitle = "Proceedings of the 2016 Conference on Empirical Methods in Natural Language Processing",
    year = "2016",
    publisher = "Association for Computational Linguistics",
    url = "https://aclanthology.org/D16-1264/",
    pages = "2383--2392"
}

@inproceedings{triviaqa,
    title = "{T}rivia{QA}: A Large Scale Distantly Supervised Challenge Dataset for Reading Comprehension",
    author = "Joshi, Mandar  and
      Choi, Eunsol  and
      Weld, Daniel S. and
      Zettlemoyer, Luke",
    booktitle = "Proceedings of the 55th Annual Meeting of the Association for Computational Linguistics (Volume 1: Long Papers)",
    year = "2017",
    url = "https://aclanthology.org/P17-1147/",
    pages = "1601--1611"
}

@inproceedings{xiao2024c,
  title={C-pack: Packed resources for general chinese embeddings},
  author={Xiao, Shitao and Liu, Zheng and Zhang, Peitian and Muennighoff, Niklas and Lian, Defu and Nie, Jian-Yun},
  booktitle={Proceedings of the 47th international ACM SIGIR conference on research and development in information retrieval},
  pages={641--649},
  year={2024}
}

@inproceedings{mgh,
    title = "Negative Matters: Multi-Granularity Hard-Negative Synthesis and Anchor-Token-Aware Pooling for Enhanced Text Embeddings",
    author = "Pan, Tengyu  and
      Duan, Zhichao  and
      Li, Zhenyu  and
      Dong, Bowen  and
      Liu, Ning  and
      Li, Xiuxing  and
      Wang, Jianyong",
    booktitle = "Proceedings of the 63rd Annual Meeting of the Association for Computational Linguistics (Volume 1: Long Papers)",
    year = "2025",
    url = "https://aclanthology.org/2025.acl-long.1501/",
    pages = "31102--31118",
}

@inproceedings{sbert,
    title = "Sentence-{BERT}: Sentence Embeddings using {S}iamese {BERT}-Networks",
    author = "Reimers, Nils  and
      Gurevych, Iryna",
    booktitle = "Proceedings of the 2019 Conference on Empirical Methods in Natural Language Processing and the 9th International Joint Conference on Natural Language Processing (EMNLP-IJCNLP)",
    year = "2019",
    address = "Hong Kong, China",
    publisher = "Association for Computational Linguistics",
    pages = "3982--3992",
}

@article{anchor,
  title={Training LLMs to be Better Text Embedders through Bidirectional Reconstruction},
  author={Su, Chang and Shi, Dengliang and Huang, Siyuan and Du, Jintao and Meng, Changhua and Cheng, Yu and Wang, Weiqiang and Lin, Zhouhan},
  journal={arXiv preprint arXiv:2509.03020},
  year={2025}
}

@article{vectoricl,
  title={Vector-ICL: In-context Learning with Continuous Vector Representations},
  author={Zhuang, Yufan and Singh, Chandan and Liu, Liyuan and Shang, Jingbo and Gao, Jianfeng},
  journal={arXiv preprint arXiv:2410.05629},
  year={2024}
}

@inproceedings{icltaskvec,
    title = "In-Context Learning Creates Task Vectors",
    author = "Hendel, Roee  and
      Geva, Mor  and
      Globerson, Amir",
    booktitle = "Findings of the Association for Computational Linguistics: EMNLP 2023",
    month = dec,
    year = "2023",
    pages = "9318--9333",
}

@inproceedings{gpt3,
  title={Language models are few-shot learners},
  author={Brown , Tom B. and Mann, Benjamin and Ryder, Nick and Subbiah, Melanie and Kaplan, Jared and Dhariwal, Prafulla and Neelakantan, Arvind and Shyam, Pranav and Sastry, Girish and Askell, Amanda and others},
  booktitle={Advances in Neural Information Processing Systems},
  volume={33},
  pages={1877--1901},
  year={2020}
}

@inproceedings{soft1,
    title = "The Power of Scale for Parameter-Efficient Prompt Tuning",
    author = "Lester, Brian  and
      Al-Rfou, Rami  and
      Constant, Noah",
    booktitle = "Proceedings of the 2021 Conference on Empirical Methods in Natural Language Processing",
    month = nov,
    year = "2021",
    pages = "3045--3059"
}

@inproceedings{soft2,
    title = "Prefix-Tuning: Optimizing Continuous Prompts for Generation",
    author = "Li, Xiang Lisa  and
      Liang, Percy",
    booktitle = "Proceedings of the 59th Annual Meeting of the Association for Computational Linguistics and the 11th International Joint Conference on Natural Language Processing (Volume 1: Long Papers)",
    month = aug,
    year = "2021",
    pages = "4582--4597"
}

@inproceedings{bellm,
    title = "{B}e{LLM}: Backward Dependency Enhanced Large Language Model for Sentence Embeddings",
    author = "Li, Xianming  and
      Li, Jing",
    booktitle = "Proceedings of the 2024 Conference of the North American Chapter of the Association for Computational Linguistics: Human Language Technologies (Volume 1: Long Papers)",
    year = "2024",
    pages = "792--804"
}

@article{yang2025task,
  title={Task Vectors in In-Context Learning: Emergence, Formation, and Benefit},
  author={Yang, Liu and Lin, Ziqian and Lee, Kangwook and Papailiopoulos, Dimitris and Nowak, Robert},
  journal={arXiv preprint arXiv:2501.09240},
  year={2025}
}

@article{gemini,
  title={Gemini embedding: Generalizable embeddings from gemini},
  author={Lee, Jinhyuk and Chen, Feiyang and Dua, Sahil and Cer, Daniel and Shanbhogue, Madhuri and Naim, Iftekhar and {\'A}brego, Gustavo Hern{\'a}ndez and Li, Zhe and Chen, Kaifeng and Vera, Henrique Schechter and others},
  journal={arXiv preprint arXiv:2503.07891},
  year={2025}
}

@article{qwen3embed,
  title={Qwen3 Embedding: Advancing Text Embedding and Reranking Through Foundation Models},
  author={Zhang, Yanzhao and Li, Mingxin and Long, Dingkun and Zhang, Xin and Lin, Huan and Yang, Baosong and Xie, Pengjun and Yang, An and Liu, Dayiheng and Lin, Junyang and Huang, Fei and Zhou, Jingren},
  journal={arXiv preprint arXiv:2506.05176},
  year={2025}
}

@article{class,
  title={An efficient framework for learning sentence representations},
  author={Logeswaran, Lajanugen and Lee, Honglak},
  journal={arXiv preprint arXiv:1803.02893},
  year={2018}
}

@article{word1,
  title={Efficient estimation of word representations in vector space},
  author={Mikolov, Tomas and Chen, Kai and Corrado, Greg and Dean, Jeffrey},
  journal={arXiv preprint arXiv:1301.3781},
  year={2013}
}

@inproceedings{word2,
  title={Glove: Global vectors for word representation},
  author={Pennington, Jeffrey and Socher, Richard and Manning, Christopher D},
  booktitle={Proceedings of the 2014 conference on empirical methods in natural language processing (EMNLP)},
  pages={1532--1543},
  year={2014}
}

@inproceedings{cse,
    title = "{CSE}: Conceptual Sentence Embeddings based on Attention Model",
    author = "Wang, Yashen  and
      Huang, Heyan  and
      Feng, Chong  and
      Zhou, Qiang  and
      Gu, Jiahui  and
      Gao, Xiong",
    booktitle = "Proceedings of the 54th Annual Meeting of the Association for Computational Linguistics (Volume 1: Long Papers)",
    year = "2016",
    url = "https://aclanthology.org/P16-1048/",
    pages = "505--515"
}

@article{wieting2015towards,
  title={Towards universal paraphrastic sentence embeddings},
  author={Wieting, John and Bansal, Mohit and Gimpel, Kevin and Livescu, Karen},
  journal={arXiv preprint arXiv:1511.08198},
  year={2015}
}

@inproceedings{zhuang2025vectoricl,
title={Vector-{ICL}: In-context Learning with Continuous Vector Representations},
author={Yufan Zhuang and Chandan Singh and Liyuan Liu and Jingbo Shang and Jianfeng Gao},
booktitle={The Thirteenth International Conference on Learning Representations},
year={2025},
url={https://openreview.net/forum?id=xing7dDGh3}
}

@article{zhao2025kalm,
  title={Kalm-embedding-v2: Superior training techniques and data inspire A versatile embedding model},
  author={Zhao, Xinping and Hu, Xinshuo and Shan, Zifei and Huang, Shouzheng and Zhou, Yao and Zhang, Xin and Sun, Zetian and Liu, Zhenyu and Li, Dongfang and Wei, Xinyuan and others},
  journal={arXiv preprint arXiv:2506.20923},
  year={2025}
}

@inproceedings{hida-etal-2025-social,
    title = "Social Bias Evaluation for Large Language Models Requires Prompt Variations",
    author = "Hida, Rem  and
      Kaneko, Masahiro  and
      Okazaki, Naoaki",
    booktitle = "Findings of the Association for Computational Linguistics: EMNLP 2025",
    year = "2025",
    url = "https://aclanthology.org/2025.findings-emnlp.783/",
    pages = "14507--14530"
}

@inproceedings{bang-etal-2025-hallulens,
    title = "{H}allu{L}ens: {LLM} Hallucination Benchmark",
    author = "Bang, Yejin  and
      Ji, Ziwei  and
      Schelten, Alan  and
      Hartshorn, Anthony  and
      Fowler, Tara  and
      Zhang, Cheng  and
      Cancedda, Nicola  and
      Fung, Pascale",
    booktitle = "Proceedings of the 63rd Annual Meeting of the Association for Computational Linguistics (Volume 1: Long Papers)",
    year = "2025",
    url = "https://aclanthology.org/2025.acl-long.1176/",
    pages = "24128--24156"
}

@article{sgpt,
  title={SGPT: GPT Sentence Embeddings for Semantic Search},
  author={Muennighoff, Niklas},
  journal={arXiv preprint arXiv:2202.08904},
  year={2022}
}

@inproceedings{gtr,
  title={Large Dual Encoders Are Generalizable Retrievers},
  author={Ni, Jianmo and Qu, Chen and Lu, Jing and Dai, Zhuyun and Ábrego, Gustavo Hernández and Ma, Ji and Zhao, Vincent Y. and Luan, Yi and Hall, Keith B. and Chang, Ming-Wei and Yang, Yinfei},
  booktitle={Proceedings of the 2022 Conference on Empirical Methods in Natural Language Processing},
  pages={9844--9855},
  year={2022}
}

@article{zhang2023language,
  title={Language models are universal embedders},
  author={Zhang, Xin and Li, Zehan and Zhang, Yanzhao and Long, Dingkun and Xie, Pengjun and Zhang, Meishan and Zhang, Min},
  journal={arXiv preprint arXiv:2310.08232},
  year={2023}
}

@inproceedings{su-etal-2023-one,
    title = "One Embedder, Any Task: Instruction-Finetuned Text Embeddings",
    author = "Su, Hongjin  and
      Shi, Weijia  and
      Kasai, Jungo  and
      Wang, Yizhong  and
      Hu, Yushi  and
      Ostendorf, Mari  and
      Yih, Wen-tau  and
      Smith, Noah A.  and
      Zettlemoyer, Luke  and
      Yu, Tao",
    booktitle = "Findings of the Association for Computational Linguistics: ACL 2023",
    month = jul,
    year = "2023",
    pages = "1102--1121"
}

@inproceedings{li-li-2024-aoe,
    title = "{A}o{E}: Angle-optimized Embeddings for Semantic Textual Similarity",
    author = "Li, Xianming  and
      Li, Jing",
    booktitle = "Proceedings of the 62nd Annual Meeting of the Association for Computational Linguistics (Volume 1: Long Papers)",
    month = aug,
    year = "2024",
    pages = "1825--1839"
}

@article{lin2025causal2vec,
  title={Causal2Vec: Improving Decoder-only LLMs as Versatile Embedding Models},
  author={Lin, Ailiang and Li, Zhuoyun and Funakoshi, Kotaro and Okumura, Manabu},
  journal={arXiv preprint arXiv:2507.23386},
  year={2025}
}

@inproceedings{sink,
  title={Look both ways and no sink: Converting llms into text encoders without training},
  author={Lin, Ziyong and Wu, Haoyi and Wang, Shu and Tu, Kewei and Zheng, Zilong and Jia, Zixia},
  booktitle={Proceedings of the 63rd Annual Meeting of the Association for Computational Linguistics (Volume 1: Long Papers)},
  pages={22839--22853},
  year={2025}
}

@inproceedings{instrut2,
    title = "One Embedder, Any Task: Instruction-Finetuned Text Embeddings",
    author = "Su, Hongjin  and
      Shi, Weijia  and
      Kasai, Jungo  and
      Wang, Yizhong  and
      Hu, Yushi  and
      Ostendorf, Mari  and
      Yih, Wen-tau  and
      Smith, Noah A.  and
      Zettlemoyer, Luke  and
      Yu, Tao",
    booktitle = "Findings of the Association for Computational Linguistics: ACL 2023",
    year = "2023",
    pages = "1102--1121"
}

@article{instrut3,
  title={Multilingual e5 text embeddings: A technical report},
  author={Wang, Liang and Yang, Nan and Huang, Xiaolong and Yang, Linjun and Majumder, Rangan and Wei, Furu},
  journal={arXiv preprint arXiv:2402.05672},
  year={2024}
}

@inproceedings{qa,
  title={Dense Passage Retrieval for Open-Domain Question Answering.},
  author={Karpukhin, Vladimir and Oguz, Barlas and Min, Sewon and Lewis, Patrick and Wu, Ledell and Edunov, Sergey and Chen, Danqi and Yih, Wen-tau},
  booktitle={Proceedings of the 2020 Conference on Empirical Methods in Natural Language Processing (EMNLP)},
  pages={6769--6781},
  year={2020}
}

@article{cmedteb,
  title   = {CMedTEB \& CARE: Benchmarking and Enabling Efficient Chinese Medical Retrieval via Asymmetric Encoders},
  author  = {Jiang, Angqing and Chen, Jianlyu and Fang, Zhe and Wang, Yongcan and Li, Xinpeng and Ding, Keyu and Lian, Defu},
  journal = {arXiv preprint arXiv:2604.10937},
  year    = {2026},
}

\appendix

\section{Experimental Details for Training}
\subsection{Training Setup}
\label{appendix:training_detail}
In this section, we provide additional training details based on Section~\ref{sect:experiment}. We fine-tune Mistral-7B for 1000 steps, and Qwen2.5-7B as well as LLaMA-3.1-8B for 800 steps. We adopt the AdamW optimizer with 300 warm-up steps, followed by a linear learning-rate decay over the remaining steps. To ensure fair comparison, we follow the open-source implementation of LLM2Vec~\cite{llm2vec} and set the random seed to 42 across all experiments. To reduce GPU memory usage, we enable bfloat16 precision, FlashAttention-2~\cite{flashattention}, and gradient checkpointing. Following~\citet{mgh}, we further employ gradient accumulation of 8 to simulate a batch size of 512. Additionally, we ensure that all samples within each batch are drawn from the same dataset

\begin{table*}[ht]
\small
\centering
\resizebox{\textwidth}{!}{
\begin{tabular}{ll}
\toprule
\textbf{Dataset} & \textbf{Instruction (s)} \\
\midrule
ELI5 & Provided a user question, retrieve the highest voted answers on Reddit ELI5 forum \\
HotpotQA & Given a multi-hop question, retrieve documents that can help answer the question \\
FEVER & Given a claim, retrieve documents that support or refute the claim \\
MIRACL & Given a question, retrieve Wikipedia passages that answer the question \\
MSMARCO Passage & Given a web search query, retrieve relevant passages that answer the query \\
MSMARCO Document & Given a web search query, retrieve relevant documents that answer the query \\
NQ & Given a question, retrieve Wikipedia passages that answer the question \\
NLI & Given a premise, retrieve a hypothesis that is entailed by the premise \\
    & Retrieve semantically similar text \\
SQuAD & Retrieve Wikipedia passages that answer the question \\
TriviaQA & Retrieve Wikipedia passages that answer the question \\
QuoraDuplicates & Given a question, retrieve questions that are semantically equivalent to the given question \\
 & Find questions that have the same meaning as the input question \\
Mr. TyDi & Given a question, retrieve Wikipedia passages that answer the question \\
DuReader & Given a Chinese search query, retrieve web passages that answer the question \\
T2Ranking & Given a Chinese search query, retrieve web passages that answer the question \\
\bottomrule
\end{tabular}}
\caption{Instructions used for publicly available retrieval datasets during training.}
\label{appendix:training_instructions}
\end{table*}
\subsection{Public Retrieval Datasets}
\label{appendix:training_data}
Following~\citet{echo}, the collection of publicly available retrieval datasets used for training is distributed under the \href{https://github.com/jakespringer/echo-embeddings/blob/master/LICENSE}{Apache License 2.0} and includes the following datasets: ELI5 (sample ratio 0.1)~\citep{eli5}, HotpotQA~\citep{hotpotqa}, FEVER~\citep{fever}, MIRACL~\citep{miracl}, MS-MARCO passage ranking (sample ratio 0.5) and document ranking (sample ratio 0.2)~\citep{marco}, NQ~\citep{karpukhin2020dense}, NLI~\citep{simcse}, SQuAD~\citep{squad}, TriviaQA~\citep{triviaqa}, Quora Duplicate Questions (sample ratio 0.1)~\citep{quora}, Mr. TyDi~\citep{MrTyDi}, DuReader~\citep{dureader}, and T2Ranking (sample ratio 0.5)~\citep{t2ranking}. 

\begin{figure*}[ht]
\centering
\includegraphics[width=0.9\textwidth]{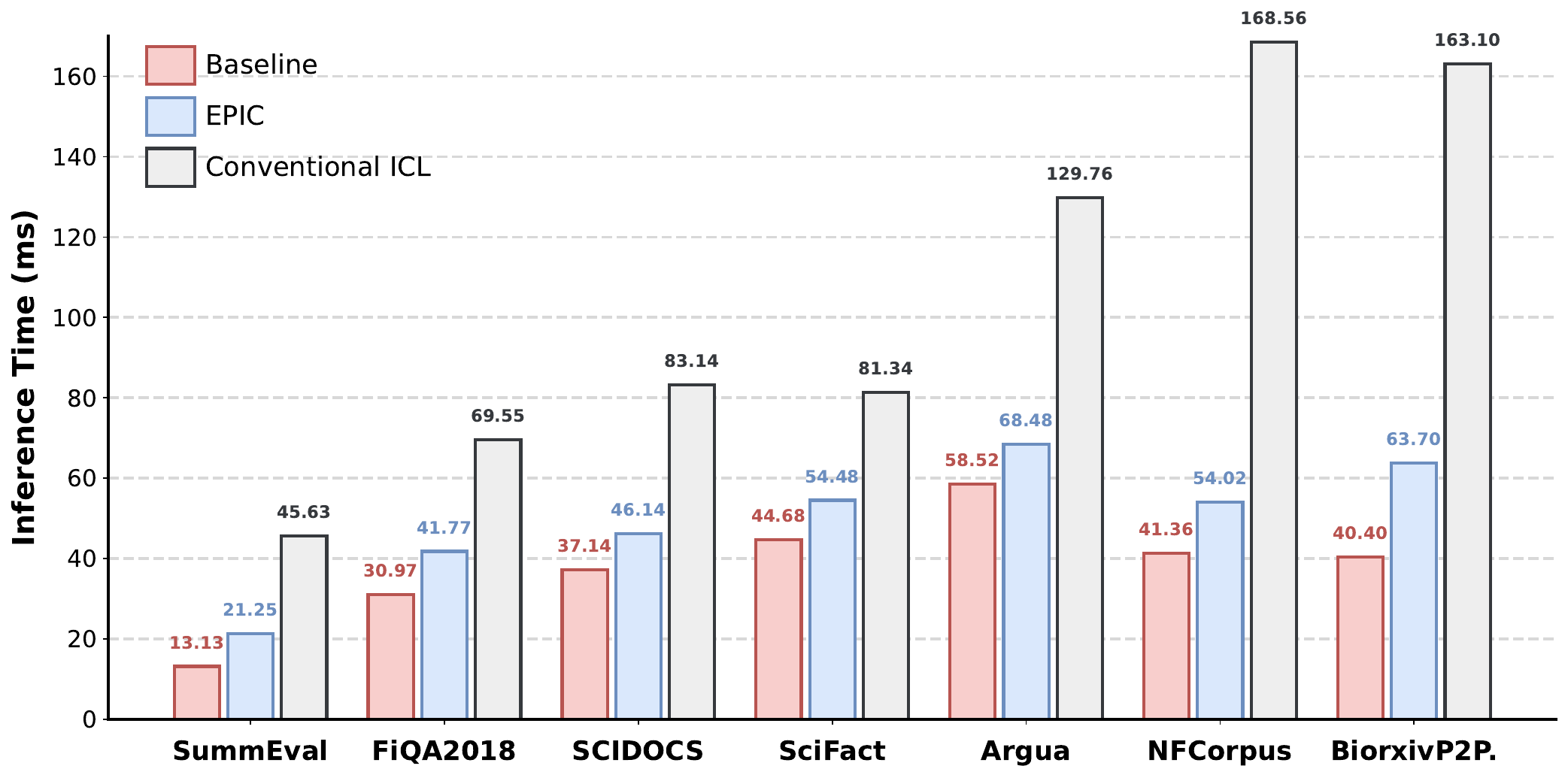}
\caption{Average per-sample inference latency of Mistral-7B–based methods on selected MTEB datasets. The baseline refers to the standard Mistral-7B model with EOS pooling. All results are obtained with a batch size of 64 on a single NVIDIA A100 GPU. For asymmetric retrieval datasets, latency is reported per \texttt{query}–\texttt{passage} pair.}
\label{fig:ifnerence_ablation}
\end{figure*}

Following~\cite{llm2vec}, we use different instructions for each retrieval dataset during training, as listed in Table~\ref{appendix:training_instructions}. It is worth noting that for \texttt{query}–\texttt{passage} sample pairs, we apply instructions only to the \texttt{query}, while leaving the \texttt{passage} unchanged.

\section{Experimental Details for Evaluation}
\label{appendix:evaluation}
\subsection{Massive Text Embeddings Benchmark (MTEB)}
In line with previous work~\cite{e5mistral,llm2vec,echo,nvembed,anchor,mgh,bgeicl}, we adopt the large-scale MTEB English subsets~\cite{mteb} to evaluate the effectiveness of our method. This benchmark is distributed under the \href{https://github.com/embeddings-benchmark/mteb/blob/main/LICENSE}{Apache License 2.0} and comprises 56 English datasets across seven diverse embedding task categories: retrieval (Retr.), reranking (Rerank.), clustering (Clust.), pair classification (PairClass.), classification (Class.), semantic textual similarity (STS), and summarization (Summ.). The corresponding evaluation metrics are nDCG@10, MAP, V-measure (V-meas.), average precision (AP), accuracy (Acc.), and Spearman correlation (Spear., both for STS and Summ.), respectively.

For fair comparison, we directly employ the in-context demonstrations curated by \href{https://github.com/FlagOpen/FlagEmbedding/tree/master/FlagEmbedding/evaluation/mteb/examples}{bge-en-icl}~\cite{bgeicl}, which provide between one and eight sentence pairs for each MTEB dataset. Since these examples are specifically selected for bge-en-icl, they may be suboptimal for our method. Therefore, for datasets where the demonstrations fail to improve performance, we simply disable in-context prompting. Notably, for asymmetric tasks such as retrieval, instructions or in-context prompts are applied only to the \texttt{query}, whereas for symmetric tasks, they are applied to both input texts. The instructions used for each MTEB dataset are listed in Table~\ref{appendix:mteb_instructions}.

\subsection{MTEB Subset}
The full MTEB benchmark contains over ten millions samples and requires hundreds of A100-80GB GPU hours to evaluate a 7B-parameter model. To accelerate ablation studies and analysis, we follow MGH~\cite{mgh} and select a representative subset of MTEB comprising 26 datasets: BIOSSES, STS12, STS13, STS14, STS15, STS16, STS17, STS22, STSBenchmark, SICK-R, AmazonReviewsClassification, MTOPDomainClassification, TweetSentimentExtractionClassification, ImdbClassification, TwitterSemEval2015, TwitterURLCorpus, SciFact, NFCorpus, FiQA2018, SCIDOCS, BiorxivClusteringS2S, MedrxivClusteringS2S, TwentyNewsgroupsClustering, AskUbuntuDupQuestions, StackOverflowDupQuestions, and SciDocsRR.

\section{Additional Results}
\subsection{Inference Latency}
In this section, we further report the inference latency of our EPIC on MTEB datasets. As shown in Figure~\ref{fig:ifnerence_ablation}, by compressing discrete textual demonstrations into embedding-based continuous representations, EPIC reduces inference time by up to 70\% compared with conventional ICL (e.g., STS22: 45.45 vs. 152.41). There findings demonstrate that our approach substantially mitigates the token burden during inference.

\subsection{The Number of In-Context Examples During Training}
By default, we randomly sample five demonstrations from the same batch during fine-tuning, following bge-en-icl~\cite{bgeicl}. We further investigate the impact of using 1, 2, and 8 demonstrations. As shown in Table~\ref{tab:ablation_number_examples}, compared to the baseline model trained without any ICL strategy, using even a single demonstration during training leads to performance improvements. However, when the number of demonstrations is increased from five to eight, the embedding performance no longer improves, while the training cost becomes higher. Overall, to ensure a fair comparison with prior work and to strike a balance between performance and computational efficiency, we use five in-context demonstrations during training in this work.
\begin{table}[t]
\centering
\scalebox{0.9}{
\begin{tabular}{c c}
\toprule
\textbf{\# of examples} & \textbf{Average Score}\\
\midrule
0  & 70.60 \\
1 & 71.15 \\
2 & 71.30 \\
\rowcolor[gray]{0.9}5 & \textbf{71.50} \\
8 & 71.43 \\
\bottomrule 
\end{tabular}
}
\caption{Performance comparison of EPIC$_\text{Mistral-7B}$ with varying numbers of in-context demonstrations during fine-tuning on the MTEB subset, where 0 examples refers to training without any ICL strategy.}
\label{tab:ablation_number_examples}
\end{table}

\subsection{Analysis of the Attention Sink Phenomenon}
\label{appendix:sink}
The attention sink phenomenon refers to the model’s tendency to focus excessively on the first token, which has been shown to hinder the performance of embedding models~\cite{sink}. We conduct an attention analysis on EPIC$_\text{Mistral-7B}$ by computing the average proportion of attention that the EOS token assigns to the first token across different layers. As shown in Table~\ref{tab:ablation_sink}, EPIC-trained models consistently alleviate the attention sink phenomenon both with and without in-context demonstrations during inference. Consequently, the EOS token, which serves as the output text embedding, can attend more effectively to the remaining tokens, thereby improving the embedding quality.
\begin{table}[h]
\centering
\scalebox{0.8}{
\begin{tabular}{l cccc}
\toprule
\textbf{Method} & \textbf{Layer 8} & \textbf{Layer 16} & \textbf{Layer 24} & \textbf{Layer 32}\\
\midrule
Baseline & \textbf{0.75} & \textbf{0.58} &\textbf{0.70} & \textbf{0.46} \\
\rowcolor[gray]{0.9}EPIC$_{\text{w/o ICD}}$ & 0.57 & 0.40 &0.47 & 0.33\\
\rowcolor[gray]{0.9}EPIC$_{\text{w/ ICD}}$ & 0.54 & 0.37 &0.45 & 0.31\\
\bottomrule 
\end{tabular}
}
\caption{Proportion of attention assigned to the first token across selected layers for EPIC-trained models with or without in-context demonstrations (ICD) during inference. Baseline models are conventionally trained without any ICL strategy.}
\label{tab:ablation_sink}
\end{table}

\begin{table}[h]
\centering
\scalebox{0.9}{
\begin{tabular}{c c}
\toprule
\textbf{LoRA Rank} & \textbf{Average Score}\\
\midrule
16 & 71.39 \\
32 & 71.42 \\
\rowcolor[gray]{0.9}64 & \textbf{71.50} \\
\bottomrule 
\end{tabular}
}
\caption{Performance comparison of EPIC$_\text{Mistral-7B}$ with different LoRA ranks on the MTEB subset.}
\label{tab:ablation_lora}
\end{table}
\subsection{The Impact of LoRA Rank}
In addition to the LoRA rank of 64 used in our experiments, we further examine the model performance with LoRA ranks of 16 and 32. As presented in Table~\ref{tab:ablation_lora}, EPIC achieves strong results even with smaller LoRA rank, demonstrating its robustness across different LoRA settings. For a fair comparison with the previous state-of-the-art method, bge-en-icl, we adopt a LoRA rank of 64 as the default setting in this work.

\label{ablation:inference}
\subsection{Detailed MTEB Results}
We present the detailed results of the proposed EPIC on all MTEB datasets using three base models: Qwen2.5-7B, Mistral-7B, and LLaMA-3.1-8B, as summarized in Table~\ref{appendix:full-mteb-results}.

\begin{table*}[h]
\centering   
\resizebox{\textwidth}{!}{
\begin{tabular}{ll}
\toprule
\textbf{Dataset} & \textbf{Instruction} \\
\midrule
AmazonCounterfactualClassification & \begin{tabular}[c]{@{}l@{}}Classify a given Amazon customer review text as either counterfactual \\ or not-counterfactual.\end{tabular} \\
AmazonPolarityClassification & Classify Amazon reviews into positive or negative sentiment  \\
AmazonReviewsClassification & Classify the given Amazon review into its appropriate rating category  \\
Banking77Classification & Given a online banking query, find the corresponding intents  \\
EmotionClassification &  \begin{tabular}[c]{@{}l@{}}Classify the emotion expressed in the given Twitter message into one of the six \\ emotions: anger, fear, joy, love, sadness, and surprise.\end{tabular}   \\
ImdbClassification & Classify the sentiment expressed in the given movie review text from the IMDB dataset.\\
MassiveIntentClassification & Given a user utterance as query, find the user intents  \\
MassiveScenarioClassification & Given a user utterance as query, find the user scenarios  \\
MTOPDomainClassification & Classify the intent domain of the given utterance in task-oriented conversation  \\
MTOPIntentClassification & Classify the intent of the given utterance in task-oriented conversation  \\
ToxicConversationsClassif. & Classify the given comments as either toxic or not toxic  \\
TweetSentimentClassification & Classify the sentiment of a given tweet as either positive, negative, or neutral  \\
ArxivClusteringP2P & Identify the main and secondary category of Arxiv papers based on the titles and abstracts. \\
ArxivClusteringS2S & Identify the main and secondary category of Arxiv papers based on the titles  \\
BiorxivClusteringP2P & Identify the main category of Biorxiv papers based on the titles and abstracts  \\
BiorxivClusteringS2S & Identify the main category of Biorxiv papers based on the titles  \\
MedrxivClusteringP2P & Identify the main category of Medrxiv papers based on the titles and abstracts  \\
MedrxivClusteringS2S & Identify the main category of Medrxiv papers based on the titles  \\
RedditClustering & Identify the topic or theme of Reddit posts based on the titles  \\
RedditClusteringP2P & Identify the topic or theme of Reddit posts based on the titles and posts  \\
StackExchangeClustering & Identify the topic or theme of StackExchange posts based on the titles  \\
StackExchangeClusteringP2P & Identify the topic or theme of StackExchange posts based on the given paragraphs  \\
TwentyNewsgroupsClustering & Identify the topic or theme of the given news articles  \\
SprintDuplicateQuestions & Retrieve duplicate questions from Sprint forum  \\
TwitterSemEval2015 & Retrieve tweets that are semantically similar to the given tweet  \\
TwitterURLCorpus & Retrieve tweets that are semantically similar to the given tweet  \\
AskUbuntuDupQuestions & Retrieve duplicate questions from AskUbuntu forum  \\
MindSmallReranking & Retrieve relevant news articles based on user browsing history  \\
SciDocsRR & Given a title of a scientific paper, retrieve the titles of other relevant papers  \\
StackOverflowDupQuestions & Retrieve duplicate questions from StackOverflow forum  \\
ArguAna & Given a claim, find documents that refute the claim  \\
ClimateFEVER & Given a claim about climate change, retrieve documents that support or refute the claim. \\
CQADupstackRetrieval &  \begin{tabular}[c]{@{}l@{}}Given a question, retrieve detailed question descriptions from Stackexchange that are \\ duplicates to the given question.\end{tabular} \\
DBPedia & Given a query, retrieve relevant entity descriptions from DBPedia  \\
FEVER & Given a claim, retrieve documents that support or refute the claim  \\
FiQA2018 & Given a financial question, retrieve user replies that best answer the question  \\
HotpotQA & Given a multi-hop question, retrieve documents that can help answer the question  \\
MSMARCO & Given a web search query, retrieve relevant passages that answer the query  \\
NFCorpus & Given a question, retrieve relevant documents that best answer the question  \\
NQ & Given a question, retrieve Wikipedia passages that answer the question  \\
QuoraRetrieval & Given a question, retrieve questions that are semantically equivalent to the given question. \\
SCIDOCS & Given a scientific paper title, retrieve paper abstracts that are cited by the given paper  \\
SciFact & Given a scientific claim, retrieve documents that support or refute the claim  \\
Touche2020 & Given a question, retrieve detailed and persuasive arguments that answer the question  \\
TRECCOVID & Given a query on COVID-19, retrieve documents that answer the query  \\
STS* & Retrieve semantically similar text.  \\
SummEval & Given a news summary, retrieve other semantically similar summaries  \\
\bottomrule
\end{tabular}}
\caption{Instructions used for MTEB evaluation. ``STS*'' indicates that the instruction is applied to all STS datasets.}
\label{appendix:mteb_instructions}
\end{table*}
\begin{table*}[t]
    \centering
    \small
    \begin{tabular}{l|ccc}
    \toprule
    \textbf{Dataset} & \textbf{Qwen2.5-7B}  & \textbf{Mistral-7B} & \textbf{LLaMA-3.1-8B} \\
    \midrule
    AmazonCounterfactualClassification & 70.55 & 77.54 & 76.87 \\
    AmazonPolarityClassification & 96.13 & 95.76 & 94.76 \\
    AmazonReviewsClassification & 54.56 & 53.91 & 51.58 \\
    ArguAna & 61.37 & 60.65 & 61.76 \\
    ArxivClusteringP2P & 51.09 & 49.69 & 49.14 \\
    ArxivClusteringS2S & 47.71 & 46.28 & 45.74 \\
    AskUbuntuDupQuestions & 64.55 & 65.96 & 64.86 \\
    BIOSSES & 86.45 & 85.99 & 86.12 \\
    Banking77Classification & 87.88 & 88.71 & 88.59 \\
    BiorxivClusteringP2P & 41.19 & 39.41 & 40.67 \\
    BiorxivClusteringS2S & 38.66 & 38.04 & 38.52 \\
    CQADupstackRetrieval & 47.13 & 44.84 & 46.95 \\
    ClimateFEVER & 35.12 & 34.01 & 35.48 \\
    DBPedia & 49.81 & 50.77 & 50.60 \\
    EmotionClassification & 51.02 & 50.49 & 49.48 \\
    FEVER & 90.55 & 91.47 & 90.90 \\
    FiQA2018 & 51.91 & 54.64 & 53.09 \\
    HotpotQA & 67.78 & 73.06 & 72.72 \\
    ImdbClassification & 93.74 & 92.26 & 92.80 \\
    MSMARCO & 40.82 & 42.04 & 42.13 \\
    MTOPDomainClassification & 96.43 & 96.04 & 96.32 \\
    MTOPIntentClassification & 82.90 & 84.18 & 85.94 \\
    MassiveIntentClassification & 79.87 & 79.23 & 78.95 \\
    MassiveScenarioClassification & 81.95 & 82.10 & 81.30 \\
    MedrxivClusteringP2P & 35.70 & 35.13 & 33.77 \\
    MedrxivClusteringS2S & 34.51 & 34.92 & 32.30 \\
    MindSmallReranking & 33.14 & 32.48 & 32.42 \\
    NFCorpus & 40.68 & 41.07 & 40.59 \\
    NQ & 63.83 & 65.81 & 67.18 \\
    QuoraRetrieval & 89.27 & 89.38 & 89.24 \\
    RedditClustering & 61.27 & 65.26 & 64.36 \\
    RedditClusteringP2P & 63.48 & 66.96 & 65.03 \\
    SCIDOCS & 23.84 & 22.03 & 22.87 \\
    SICK-R & 83.57 & 83.80 & 84.10 \\
    STS12 & 78.74 & 79.04 & 79.50 \\
    STS13 & 89.17 & 89.18 & 89.42 \\
    STS14 & 84.81 & 85.72 & 85.77 \\
    STS15 & 89.61 & 90.59 & 90.04 \\
    STS16 & 87.99 & 88.44 & 88.33 \\
    STS17 & 91.86 & 92.78 & 91.95 \\
    STS22 & 69.20 & 69.99 & 69.39 \\
    STSBenchmark & 88.69 & 89.38 & 89.21 \\
    SciDocsRR & 86.61 & 84.42 & 85.76 \\
    SciFact & 77.37 & 76.90 & 77.66\\
    SprintDuplicateQuestions & 96.73 & 96.58 & 96.96 \\
    StackExchangeClustering & 74.99 & 73.71 & 71.65 \\
    StackExchangeClusteringP2P & 39.38 & 38.68 & 36.38 \\
    StackOverflowDupQuestions & 53.81 & 55.21 & 53.82 \\
    SummEval & 30.86 & 31.41 & 31.26 \\
    TRECCOVID & 83.60 & 83.10 & 79.65 \\
    Touche2020 & 24.70 & 23.56 & 25.32 \\
    ToxicConversationsClassification & 61.34 & 65.20 & 65.55 \\
    TweetSentimentExtractionClassification & 63.58 & 62.25 & 62.17 \\
    TwentyNewsgroupsClustering & 55.51 & 57.12 & 57.81 \\
    TwitterSemEval2015 & 79.94 & 82.13 & 80.02 \\
    TwitterURLCorpus & 87.28 & 87.16 & 86.97 \\
    \midrule
    \textbf{MTEB Average (56)} & 65.97 & 66.37 & 66.10 \\
    \bottomrule
    \end{tabular}
    \caption{Results of EPIC on each MTEB datasets across three base models: Qwen2.5-7B, Mistral-7B, and LLaMA-3.1-8B.}
    \label{appendix:full-mteb-results}
\end{table*}

\end{document}